\RequirePackage{amsthm}

\documentclass[pdflatex,sn-vancouver,Numbered]{sn-jnl}

\usepackage{graphicx}%
\usepackage{multirow}%
\usepackage{amsmath,amssymb,amsfonts}%
\usepackage{amsthm}%
\usepackage{mathrsfs}%
\usepackage[title]{appendix}%
\usepackage{xcolor}%
\usepackage{textcomp}%
\usepackage{manyfoot}%
\usepackage{booktabs}%
\usepackage[linesnumbered, ruled, vlined]{algorithm2e}
\usepackage{algpseudocode}%
\usepackage{listings}%
\usepackage{enumitem}
\usepackage[caption=false]{subfig}%
\usepackage{lmodern}

\theoremstyle{thmstyleone}%
\theoremstyle{thmstyletwo}%

\theoremstyle{thmstylethree}%

\begin{document}

\title[Kolmogorov–Arnold Networks-based GRU and LSTM for Loan Default Early Prediction]{Kolmogorov–Arnold Networks-based GRU and LSTM for Loan Default Early Prediction}


\author[1]{\fnm{Yue} \sur{Yang}}\email{scxyy2@nottingham.edu.cn}

\author[2]{\fnm{Zihan} \sur{Su}}\email{smyzs4@nottingham.edu.cn}

\author[2]{\fnm{Ying} \sur{Zhang}}\email{smyyz22@nottingham.edu.cn}

\author[1]{\fnm{Chang Chuan} \sur{Goh}}\email{chang.goh@nottingham.edu.cn}

\author[1]{\fnm{Yuxiang} \sur{Lin}}\email{ssyyl35@nottingham.edu.cn}

\author*[1]{\fnm{Anthony Graham} \sur{Bellotti}}\email{anthony-graham.bellotti@nottingham.edu.cn}

\author*[1]{\fnm{Boon Giin} \sur{Lee}}\email{boon-giin.lee@nottingham.edu.cn}


\affil[1]{\orgdiv{School of Computer Science}, \orgname{University of Nottingham Ningbo China}, \city{Ningbo}, \postcode{315100}, \state{Zhejiang}, \country{China}}

\affil[2]{\orgdiv{Department of Mathematical Sciences}, \orgname{University of Nottingham Ningbo China}, \city{Ningbo}, \postcode{315100}, \state{Zhejiang}, \country{China}}


\abstract{This study addresses a critical challenge in time series anomaly detection: enhancing the predictive capability of loan default models more than three months in advance to enable early identification of default events, helping financial institutions implement preventive measures before risk events materialize. Existing methods have significant drawbacks, such as their lack of accuracy in early predictions and their dependence on training and testing within the same year and specific time frames. These issues limit their practical use, particularly with out-of-time data. To address these, the study introduces two innovative architectures, GRU-KAN and LSTM-KAN, which merge Kolmogorov–Arnold Networks (KAN) with Gated Recurrent Units (GRU) and Long Short-Term Memory (LSTM) networks. The proposed models were evaluated against the baseline models (LSTM, GRU, LSTM-Attention, and LSTM-Transformer) in terms of accuracy, precision, recall, F1 and AUC in different lengths of feature window, sample sizes, and early prediction intervals. The results demonstrate that the proposed model achieves a prediction accuracy of over 92\% three months in advance and over 88\% eight months in advance, significantly outperforming existing baselines.}




\maketitle

\section{Introduction}

Time series anomaly detection is important in data mining and machine learning, focused on identifying unusual patterns in sequential data. These anomalies often indicate considerable deviations from normal behaviors and signify potential risks. In finance, predicting loan default is a key application. Loan default refers to when a borrower cannot repay the loan principal and interest on time or if the loan is extensively overdue. Predicting loan default involves analyzing extensive historical data and borrower traits using data analysis and machine learning to predict potential defaults. This prediction is critical for financial institutions, enabling them to identify high-risk borrowers and take preventive steps to mitigate non-performing loans, ultimately enhancing asset quality and financial health.

Although existing loan default prediction models such as LSTM can, to some extent, identify high-risk borrowers, they still face limitations in predictive accuracy and practical use. Typically, financial models such as those based on LSTM struggle to deliver accurate, early predictions, depriving financial institutions of the time needed to prepare and respond. Early predictions of defaults are crucial for banks to implement risk reduction strategies and minimize potential losses. Timely forecasts allow banks to revise lending strategies, improve risk management by monitoring processes, and even engage with borrowers to assess their repayment capacity. If predictions occur only at the onset of default, the practical utility of the model is significantly reduced.

Existing models often depend on training and testing with out-of-sample (OOS) data, meaning that both sets are of the same year. This approach affects the ability of the model to quickly integrate new data, causing the model's prediction results to not reflect the current risk conditions. Researchers face the challenge of waiting until enough data accumulate, which compromises its real-time applicability. The constantly changing financial environment requires banks to update their data analysis. This constraint not only extends the model development timeline, but also restricts the training data, impairing the utilization of extensive historical data.

The novelty of this study lies in its aim to overcome the limitations of existing anomaly detection models in accurately predicting loan defaults more than three months in advance, providing financial institutions with sufficient time for proactive risk management. Specifically, we investigate designing a model capable of training and testing in different years, improving early prediction over current models. We introduce two innovative models, LSTM-KAN and GRU-KAN, based on Kolmogorov–Arnold Networks (KAN), a leading deep learning framework. We evaluated these models in three experimental scenarios. The first scenario assesses each effectiveness of models with various feature window lengths, indicating the number of historical months used for training, to determine whether similar or better predictive accuracy is achievable with shorter windows compared to baseline models. The second scenario tests the impact of a blank interval between the input feature window and the observation period, simulating the early prediction of defaults. The third scenario examines the efficiency of models across different sample sizes, with the aim of showing equal or better performance even with smaller datasets. All scenarios use out-of-time (OOT) test sets, which means that training and testing data are from different years. This setup demonstrates the ability of the proposed models to utilize extensive historical data, showing adaptability to new data and potential for near-real-time applications.

The primary contributions of our study are:
\begin{enumerate}
    \item To introduce innovative KAN-based GRU and LSTM models that flexibly optimize activation functions for adaptable modeling of complex nonlinear relationships in time series data. This method improves predictive precision and robustness in forecasting loan defaults.
    
    \item To introduce the inclusion of a ``blank interval'' enables the development of a model for the early prediction of loan defaults, serving as a key tool for financial institutions to manage risk proactively. Early identification of default risks allows banks to take preventive actions, modify credit policies, and mitigate financial losses.
    
    \item To integrate the Transformer architecture into loan default prediction and performing comparative experiments to assess its ability to capture complex patterns and long-term dependencies addresses a gap in current research.
    
    \item To propose models that demonstrate enhanced performance in handling OOT data under various conditions, as validated by experimental findings. This feature improves its suitability for near-real-time anomaly detection by integrating the most recent data in dynamic financial contexts.
    
\end{enumerate}

This study aims to improve the prediction of loan default by overcoming the shortcomings of current models and increasing their predictive power. We aim to provide financial institutions with a robust solution for identifying and mitigating loan default risks through an innovative model design and comprehensive evaluation.

\section{Related Works}

\subsection{Early Prediction with OOT data}

Previous studies on predicting loan defaults show differences and limitations in aspects such as test set construction and utilization of time series data. For example, \citet{2024_9} used loan repayment data from loans issued in 2009 and 2010 in the Freddie
Mac dataset, covering the period from January 2012 to June 2013, with a feature window every six months, totaling 13 windows for the training set; and used the repayment data for July to December 2013 as the feature window for the test set to forecast defaults during the next 12 months. Being a test set derived from different segments of the same borrowers' repayment behavior time series as the training set, the patterns in the test data resemble those of the training set, potentially leading to overly optimistic prediction results. Furthermore, the prolonged time after loan issuance to attain the test set indicates that the study may not have fully addressed the practical applicability of the model.

\citet{2024_10} integrated a survival analysis model with LSTM to examine variations in default rates during the life of the loan. By using a survival model where defaults are noted at a specific loan age, assessing default risk within the observation period post-issuance was unnecessary as is required in static credit risk models. The study uses an OOS test set from 2004 to 2013, which matches the training set in length and year. However, this data-conformity method imposes limitations. Using OOS data from the same year for both training and testing reduces the capability of the model to utilize a broad historical dataset while incorporating recent information. This restricts its effective use in real-time, with sufficient data available only after delays. This limitation highlights the necessity of OOT data, which better facilitates real-time applications.



\citet{2023_30} investigated the integration of machine learning with survival analysis to predict loan defaults. Using panel data from 1999 to 2019, the study used both OOS and OOT test sets over a 24-month observation period. However, the length of this period affected the model in accurately identifying if a default might occur sooner, such as within three months, thus limiting its usefulness in time-sensitive predictions. \citet{2023_36} examined loan default prediction but omitted details on handling time series data and test set design, viewing it as a static binary classification issue for unbalanced data. Similarly, \citet{2020_3} developed the HUlHEN model to better profile loan applicants, improving the accuracy of default prediction, but also neglected time series data handling and experimental design. \citet{2020_15}, \citet{2019_5}, and \citet{2019_8} also reduced the problem to a static binary classification, not adequately addressing temporal factors.

In summary, existing studies on loan default prediction have mainly used static data and ignored the ability of time series data to detect temporal patterns and trends. Even studies using time series data have mostly concentrated on OOS data, with OOT data being less utilized. Furthermore, few have attempted to integrate blank intervals to simulate ``early prediction''. These limitations have led to a poor understanding of the early prediction performance of existing models and a lack of exploration of the dynamic nature of time series data. To address these issues, our study aims to develop models using OOT data that excel in early prediction tasks, thereby improving the accuracy and practical applicability of loan default prediction.


\subsection{Time Series Anomaly Detection}

GRU and LSTM are commonly used in time series research \citep{2021_13}, serving as a solid basis for our proposed improvements. This section reviews research utilizing LSTM and GRU algorithms, with LSTM often preferred for time series due to its ability to capture long-term dependencies and contextual information.

\subsubsection{GRU related applications}

\citet{2020_7} assessed the GRU model against LSTM, ARIMA, SVM, and ANN through two cross-validation techniques and three temporal dimensions using monthly loan default rates from the Lending Club P2P platform (2008-2015) \citep{LendingClub}. The findings indicated better performance of LSTM in several metrics, while GRU effectively handles long-term dependencies with reduced computational demand, making it suitable for time series analysis. \citet{2024_21} compared the prediction accuracy of five machine learning classifiers (Gaussian Naive Bayes, AdaBoost, Gradient Boosting, K-Nearest Neighbors, Decision Trees, Random Forest and Logistic Regression) and eight deep learning algorithms (multi-layer perceptron (MLP), convolutional neural networks (CNN), LSTM, Transformer, GRU, Autoencoder, ResNet, and DenseNet) to forecast loan defaults using four metrics.

\citet{2023_47} presented the LSTM-GRU model for credit scoring, which integrates GRU and LSTM. Experiments revealed that this model outperformed baseline models such as SVM, CNN, LSTM, and GRU in precision, recall, F1 score, and AUC. The study highlighted the importance of choosing the right deep learning model for credit scoring due to variations in performance and prediction ability. \citet{2023_45} introduced GRN, an interpretable multivariate time series anomaly detection method utilizing Graph Neural Networks and GRU. GRU was selected for its proficiency in capturing long- and short-term dependencies in time series data with a lower computational cost than LSTM. \citet{2019_10} developed a novel GRU-based anomaly detection approach for network logs, using the support vector domain description (SVDD). Experiments with the KDD Cup99 dataset indicated that it outperformed the classical GRU-MLP and LSTM methods. Similarly, \citet{2018_5} designed GGM-VAE, a GRU-based Gaussian mixture anomaly detection system, integrating GRU with Gaussian mixture priors to model multimodal data distributions. The tests showed significant improvements in accuracy and F1 scores over traditional GRU models. These investigations highlighted the effectiveness of GRU-based designs in improving time series anomaly detection.

\subsubsection{LSTM related applications}

\citet{2021_11} utilized LSTM as a model for detecting credit card fraud, focusing on historical purchase patterns to enhance the detection accuracy of new fraudulent transactions. \citet{2023_25} created a rural microcredit risk assessment model based on LSTM, named the self-organizing LSTM algorithm, designed to capture the long-term dependencies between past actions and associated risk factors of borrowers. \citet{2023_26} introduced the sliding window and the attention mechanism LSTM (SL-ALSTM), which combines LSTM with a sliding window and the attention mechanism. This model maximizes ability of LSTM to capture both long-term and short-term dependencies in time series, using sliding windows for extracting context from loan data and attention mechanism for weighting crucial data, thus improving its ability to identify patterns and improve time series prediction. On the Lending Club public dataset, it surpassed ARIMA, SVM, ANN, LSTM, and GRU models. 

\citet{2019_7} proposed a similar architecture of attention-based LSTM (AT-LSTM) for the prediction of financial time series, which also uses attention to select relevant input features for the LSTM network. \citet{2021_5} developed a credit card fraud detection system using data sequence modeling, LSTM, and an attention mechanism, which assigns varying attention to information from the hidden layer of LSTM. This system considers the sequential structure of transactions, identifying crucial transactions in the input and improving the prediction of fraudulent activities. Experiments revealed that this model achieved the highest precision and recall compared to base models such as LSTM, GRU, and ANN. \citet{2019_6} introduced LGMAD, a real-time anomaly detection algorithm  which was validated on the NAB dataset \citep{NABDataset} utilizing LSTM and a Gaussian mixture model (GMM). The anomalies in each univariate sensor time series are assessed with the LSTM, followed by multidimensional anomaly detection with the GMM.


In recent years, the emergence of Transformers has led to numerous studies exploring their use for time series data prediction \citep{2022_17}. However, opinions differ among scholars about the suitability of Transformers for these tasks. \citet{2023_6} used Transformer models for innovating time series anomaly detection by simultaneously assessing global and local patterns. The model features an encoder with multiple Transformer layers and a decoder with one-dimensional convolutional layers; anomaly scores are the deviations between predicted and actual values at each timestamp. Benchmark tests show that the proposed approach exceeds the performance of LSTM and CNN, underscoring the effectiveness of the Transformer layers. \citet{LSTM-transformer} introduced a forecasting model that merges the Transformer self-attention mechanism with LSTM to efficiently capture long-term dependencies. To improve search efficiency, a hybrid of random search and Bayesian optimization refines parameters and regularization. The accuracy of the LSTM-Transformer model is validated by comparisons with LSTM, CNN, Transformer, and CNN-LSTM, showing that it achieves the best accuracy. \citet{2023_9} linked BiLSTM to the three-layer Transformer encoder for return prediction to increase the performance of the portfolio model. The BiLSTM-Transformer model initially predicted alternative asset returns, which were then incorporated
into a mean-variance (MV) model. Six constituent stocks of the US30 index
were used as alternative assets for 270 investments, and comparisons with the LSTM and Transformer models validated the performance of BiLSTM-Transformer.

In contrast, \citet{2022_17} questioned the effectiveness of popular Transformer-based methods in long-term time series forecasting (LTSF). The study used a straightforward linear model, LTSF-Linear, as the direct multi-step (DMS) benchmark to support this assertion. Importantly, the key contribution of the study is not the linear model itself but the identification of a significant concern, as shown through comparative experiments, indicating that Transformers might not be universally effective as claimed. Meanwhile, \citet{2025_1} evaluated 75 models in 16 datasets, showing that model performance varies by dataset and that both machine learning and deep learning models are dataset-specific. Hence, the critique of Transformers does not definitively prove that integrating Transformer layers with any model would fail on all time series datasets. Since Transformers have not been extensively applied to loan default prediction in the context of time series anomaly detection, exploring Transformer architectures in this area is still valuable to assess if they improve the capture of intricate patterns and long-term dependencies, thus addressing current research limitations.


\subsection{KAN related applications}

KAN, pioneered by an MIT research team \citep{2024_17}, is an innovative model that can potentially be used for detecting anomalies in time series. The method decomposes complex time series into univariate functions, replacing traditional linear weights with spline-parameterized alternatives, thus enabling dynamic learning of activation patterns and significantly enhancing interpretability. \citet{2024_15} introduced KAN-AD to address KAN’s susceptibility to local anomalies caused by optimizing univariate functions with spline functions. KAN-AD emphasizes global patterns with Fourier series, reducing the impact of local peaks. Consequently, by transforming existing black-box learning methods into learning weights for univariate functions, it improves both effectiveness and efficiency.

\citet{2024_16} investigated how KAN can be used for time series forecasting, introducing T-KAN and MT-KAN. T-KAN is oriented to detect concept drift, using symbolic regression to explain non-linear links between predictions and previous time steps, ensuring interpretability in dynamic contexts. MT-KAN improves forecasting by extracting and using complex connections between variables in multivariate time series. Experiments validated the efficacy of these methods, with T-KAN and MT-KAN significantly outperforming baseline models such as MLP, RNN, and LSTM in forecasting tasks. \citet{2024_19} explored a the use of KAN in time series prediction using its adaptive activation functions to improve predictive modeling. The experiments showed that KAN surpassed traditional MLP in real-world satellite traffic forecasting with fewer learnable parameters and higher accuracy. \citet{2024_20} proposed a forecasting model called C-KAN for multistep predictions, where it combines convolutional layers with the KAN architecture. This model uses convolutional layers to capture the behavior and internal patterns of time series data, facilitating promising feature analysis and potentially more precise input-output management.

KAN, considered an advanced MLP variant, has not yet been studied for the prediction of loan default. The originality and potential of such innovative algorithms in offering new insights justify exploring the application and effectiveness of KAN in this area, thus expanding its usage.


\section{Methodology}
\subsection{Design of GRU-KAN and LSTM-KAN}

This study explores the interaction between different feature extraction methods and the KAN model, proposing a hybrid architecture that integrates GRU-KAN and LSTM-KAN to enhance the ability to capture complex patterns in time series data. Specifically, both models share a similar architecture comprising a data preprocessing layer, a masking layer, a feature extraction layer, a KAN layer, a fully connected layer, and an output layer. The key difference between the two lies in the recurrent neural network layer used: GRU-KAN employs a GRU network, while LSTM-KAN uses aLSTM network. LSTM retains long-term dependencies effectively through its gating mechanism in the feature extraction layer, whereas GRU simplifies this process with fewer parameters, achieving a lightweight model while maintaining robust temporal modeling capabilities.

\begin{figure}[!ht]
\centering
\subfloat{%
\resizebox*{13cm}{!}{\includegraphics{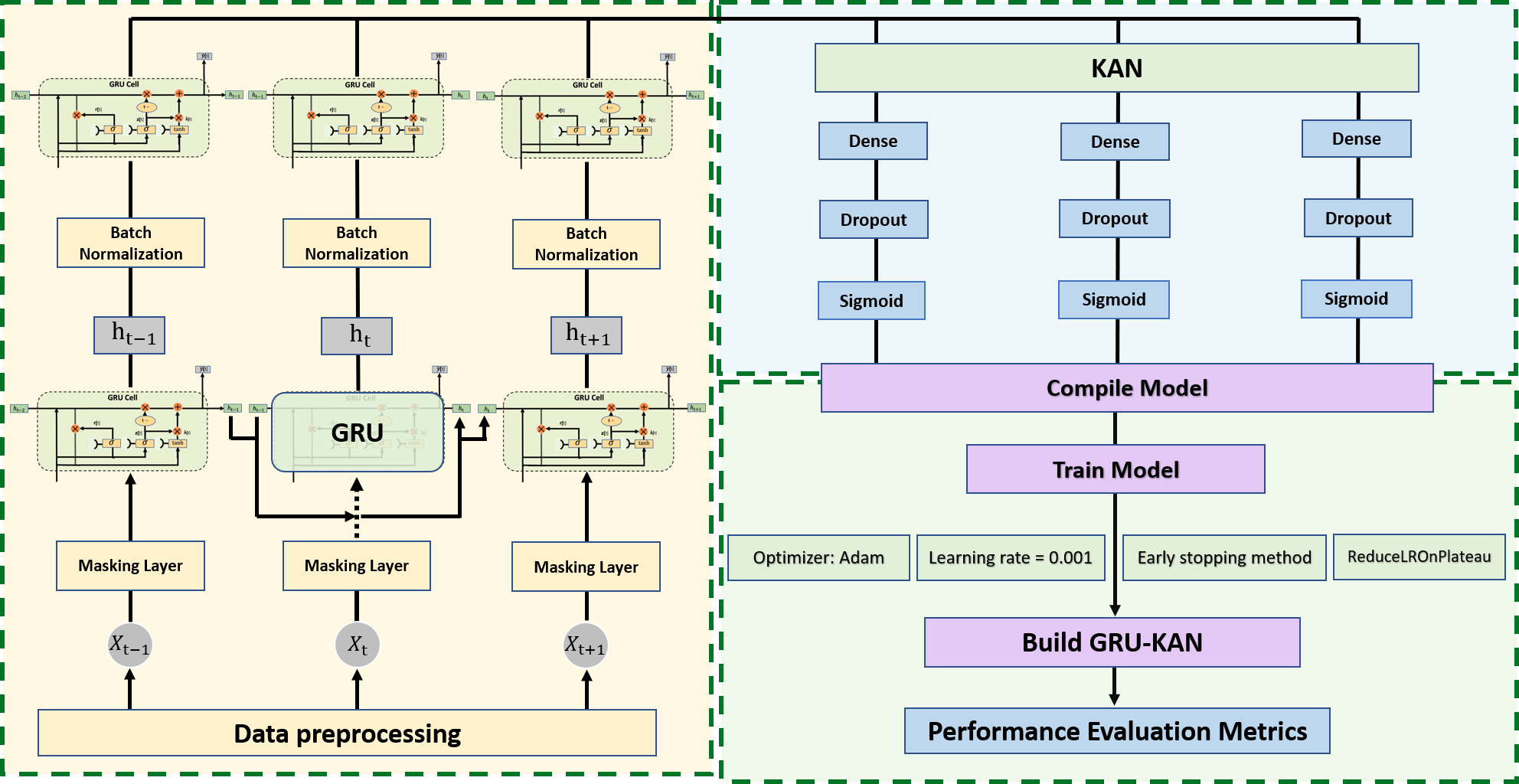}}}
\caption{Configuration of the proposed GRU-KAN.} \label{fig:GRU-KAN}
\end{figure}

\begin{figure}[!ht]
\centering
\subfloat{%
\resizebox*{13cm}{!}{\includegraphics{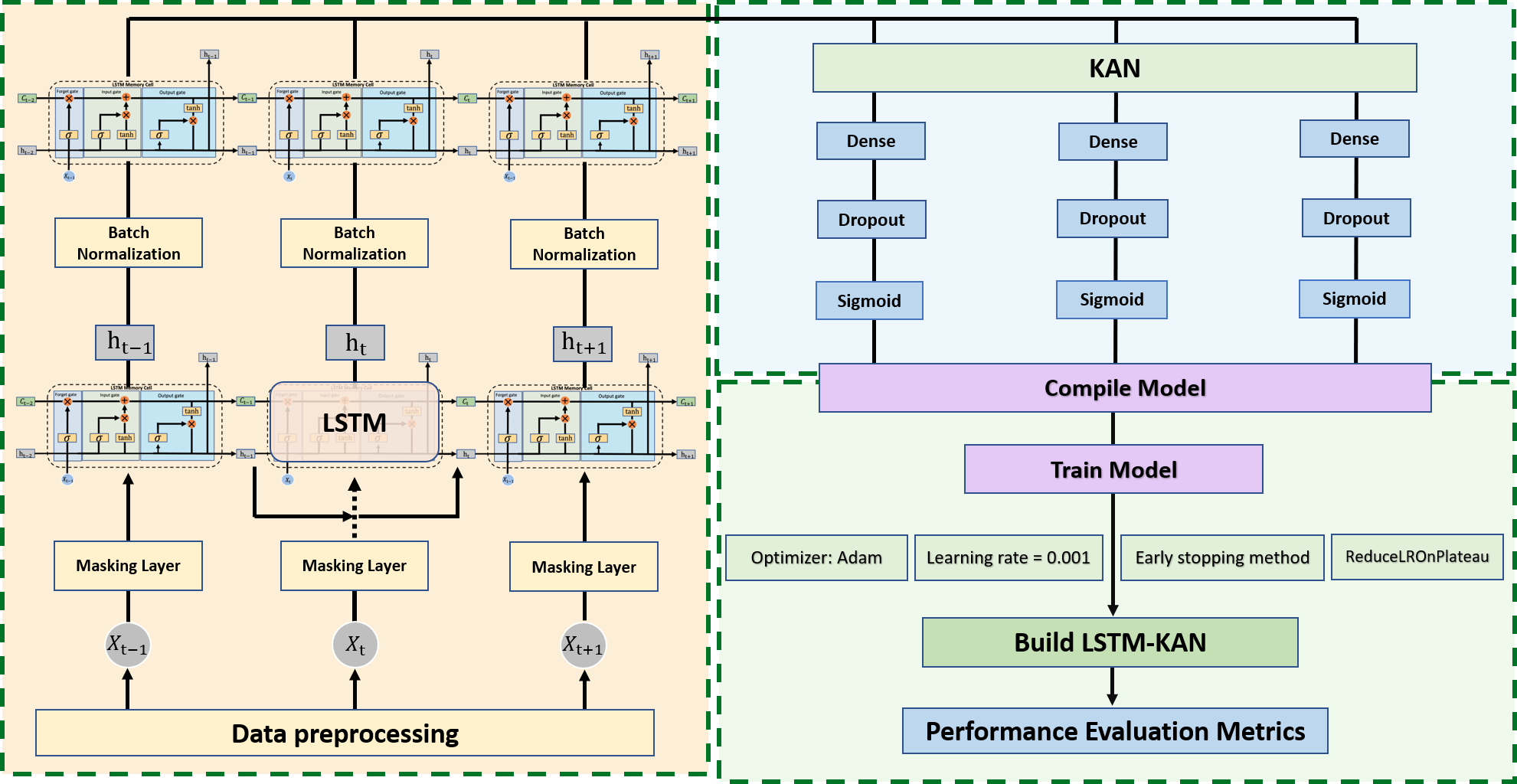}}}
\caption{Configuration of the proposed LSTM-KAN} \label{fig:LSTM-KAN}
\end{figure}

Figures \ref{fig:GRU-KAN} and \ref{fig:LSTM-KAN}, along with the pseudocode in Algorithms \ref{alg1} and \ref{alg2}, provide detailed descriptions of the GRU-KAN and LSTM-KAN architectures. The term "model" in the following definitions applies to both GRU-KAN and LSTM-KAN due to their shared structural characteristics.

\begin{algorithm}[!ht]
\caption{Pseudocode of the proposed GRU-KAN}
\label{alg1}
\small
\KwIn{Data $X_i, 1 < i < N$}
Initialize model M(GRU, KAN) with parameters $\theta$ \\
\For{\textrm{\upshape each input sequence} $X$ \textrm{\upshape from 1 to} $N$}{
$X' \gets$ Preprocess($X,\theta_{masking}$) \\
$Masked_{input} \gets$ Masking(mask\_value=0.0)($X'$) \\
$h_{gru1} \gets$ GRU(units=128, return\_sequences=True, $\theta_{gru}$)($Masked_{input}$) \\ 
$h_{gru1_{norm}} \gets$ BatchNormalization($\theta_{bn}$)($h_{gru1}$) \\
$h_{gru2} \gets$ GRU(units=64, return\_sequences=False, $\theta_{gru}$)($h_{gru1_{norm}}$) \\ 
$kan_{out} \gets$ KAN(output\_dim=1, num\_functions=10, $\theta_{kan}$)($h_{gru2}$) \\
$dense_{out} \gets$ Dense(units=64, activation=`relu', $\theta_{dense}$)($kan_{out}$) \\
$dropout_{out} \gets$ Dropout(rate=0.3, $\theta_{dropout}$)($dense_{out}$) \\
$instance_{output} \gets$ Dense(units=1, activation=`sigmoid', $\theta_{output}$)($dropout_{out}$) \\
Return $instance_{output}$
}
\KwOut{M(GRU, KAN, $\theta_{output}$), $instance_{output}$} 
\end{algorithm}

\begin{algorithm}[!ht]
\caption{Pseudocode of the proposed LSTM-KAN}
\label{alg2}
\small
\KwIn{Data $X_i, 1 < i < N$}
Initialize model M(LSTM, KAN) with parameters $\theta$ \\
\For{\textrm{\upshape each input sequence} $X$ \textrm{\upshape from 1 to} $N$}{
$X' \gets$ Preprocess($X,\theta_{masking}$) \\
$Masked_{input} \gets$ Masking(mask\_value=0.0)($X'$) \\
$h_{lstm1} \gets$ LSTM(units=128, return\_sequences=True, $\theta_{lstm}$)($Masked_{input}$) \\ 
$h_{lstm1_{norm}} \gets$ BatchNormalization($\theta_{bn}$)($h_{lstm1}$) \\
$h_{lstm2} \gets$ LSTM(units=64, return\_sequences=False, $\theta_{lstm}$)($h_{lstm1_{norm}}$) \\ 
$kan_{out} \gets$ KAN(output\_dim=1, num\_functions=10, $\theta_{kan}$)($h_{lstm2}$) \\
$dense_{out} \gets$ Dense(units=64, activation=`relu', $\theta_{dense}$)($kan_{out}$) \\
$dropout_{out} \gets$ Dropout(rate=0.3, $\theta_{dropout}$)($dense_{out}$) \\
$instance_{output} \gets$ Dense(units=1, activation=`sigmoid', $\theta_{output}$)($dropout_{out}$) \\
Return $instance_{output}$
}
\KwOut{M(LSTM, KAN, $\theta_{output}$), $instance_{output}$} 
\end{algorithm}

The data pre-processing layer is responsible for preparing loan-related input data before feeding it into the model. A detailed description of this process can be found in Section \ref{sec:3.2.2}. The masking layer identifies padding values in the input data. Since loan-related time series data often have varying sequence lengths, shorter sequences are typically padded to match the longest sequence for batch processing. The masking layer distinguishes between actual data and padding values, ensuring the model focuses only on meaningful inputs while ignoring artificial padding. This step prevents training interference from filler values, allowing the model to capture essential patterns more effectively and improve prediction accuracy.

After masking, the data is passed to the LSTM layer for feature extraction, where two different methods—LSTM and GRU—are applied and compared. Both GRU and LSTM are variants of Recurrent Neural Networks (RNNs) designed to effectively handle time-series data by capturing dependencies across different time steps. For example, in loan data, a borrower's repayment behavior at different time points is inherently correlated. By leveraging LSTM or GRU layers, the model can learn these temporal dependencies and extract dynamic features related to loan default risks.

In this study, both models employ a two-layer feature extraction structure, where the sequential data is first processed through two stacked LSTM or GRU layers. To improve training stability and generalization, a Batch Normalization layer is introduced between the two feature extraction layers. \\

\textbf{Feature Extraction Stage:}\\

LSTM (Long Short-Term Memory) is a variant of RNN (Recurrent Neural Network) that introduces a gating mechanism and memory cells to mitigate the vanishing gradient problem, thereby slowing down memory decay during back-propagation.

Totally, LSTM uses six gates to regulate the flow of information. The LSTM cell first processes information through the forget gate:
\begin{equation}
    f_t = \sigma(W_f[h_{t-1}]+b_f)
\end{equation}
where $\sigma$ is the sigmoid activation function, and $W_f$, $b_f$ are trainable parameters.

After the forget gate, the LSTM cell evaluates new information through the input gate and the candidate cell state. The input gate is defined as
\begin{equation}
    i_t = \sigma(W_i[h_{t-1},x_t]+b_i)
\end{equation}
while the candidate cell state is computed as 
\begin{equation}
    \tilde{c}_t = \tanh(W_c[h_{t-1},x_t]+b_c)
\end{equation}
The input gate regulates how much new information is added to the memory cell, while the candidate cell state represents the new memory content that could be incorporated.

The cell state is then updated as a combination of the old state, filtered by the forget gate, and the new candidate state. The update process follows the equation
\begin{equation}
    c_t = f_t \odot c_{t-1} + i_t \odot \tilde{c}_t
\end{equation}
where $\odot$ represents element-wise multiplication. This step ensures that relevant past information is preserved while new meaningful data is integrated into the memory.

Finally, the output gate determines the hidden state, which serves as both the short-term memory and the output of the LSTM cell. This is calculated using the formula
\begin{equation}
    o_t = \sigma(W_o[h_{t-1},x_t]+b_o)
\end{equation}
followed by
\begin{equation}
    h_t = o_t \odot \tanh(c_t)
\end{equation}
 The output gate decides how much of the cell state should be exposed, while the hidden state serves as the output for the current step and as input for the next step.
 
When the feature extraction layer consists of LSTM layers, it includes two key components: the first LSTM layer with 128 units and the second LSTM layer with 64 units. Assume that the masked input time series is $ Masked_{input} = \{ x_1, x_2, ... ,x_T\} $

The first LSTM layer is structured to output a sequence of hidden states $ h_{lstm1} = \{h_1, h_2,...,h_{T} \} $ corresponding to each time step, rather than a single summary representation. This design choice allows the model to retain the temporal structure of the input data, ensuring that the extracted features capture both short-term and long-term dependencies. 

Following the first LSTM layer, a Batch Normalization (BatchNorm) layer is applied to stabilize training and improve generalization. This normalization process is governed by the following formula:
\begin{equation}
    h_{norm} = \gamma\frac{h_{lstm1} - \mu}{\sigma + \epsilon} + \beta
\end{equation}

where:

\begin{itemize}
    \item $\mu$ and $\sigma$ are the mean and standard deviation computed over the batch.
    \item $\gamma$ and $\beta$ are learnable parameters.
    \item $\epsilon$ is a small constant added for numerical stability, preventing division by zero.
\end{itemize}

By adjusting the mean to zero and the variance to one, Batch Normalization effectively mitigates the issues of vanishing or exploding gradients, leading to a more stable optimization process. Additionally, it accelerates convergence, enabling faster training and improved overall performance of the model.

After passing through the first LSTM layer and the Batch Normalization layer, the data has been transformed into a stable and feature-rich representation. To further refine these extracted features, a second LSTM layer is introduced, which selectively emphasizes the most critical information through its gating mechanisms. This layer focuses on capturing the most salient temporal dependencies from the normalized sequence, ensuring that only the most relevant patterns are retained for subsequent processing.

Unlike the first LSTM layer, which preserves the full temporal structure, the second LSTM layer is designed to summarize the learned information into a single hidden state by returning only the final time step’s hidden representation. This compressed output forms the final feature vector $ h_{lstm2} \in R^{64} $, where 64 represents the number of hidden units in the second LSTM layer.

GRU (Gated Recurrent Unit) is a type of Recurrent Neural Network (RNN) that simplifies the structure of Long Short-Term Memory (LSTM) while effectively addressing the vanishing gradient problem. Unlike standard RNNs, which struggle with long-term dependencies due to gradient decay, GRU introduces two gates: the reset gate and the update gate. These gates regulate the flow of information, allowing the model to retain essential historical data while discarding irrelevant past information. Compared to LSTM, GRU has fewer parameters and is computationally more efficient while maintaining comparable performance in many sequence modeling tasks.

The update gate plays a crucial role in determining how much of the previous hidden state $ h_{t-1} $ should be carried forward to the current hidden state $ h_t $. It helps the model decide whether to retain past information or update it with new input from the current time step.

Mathematically, the update gate is computed as:
\begin{equation}
    z_t = \sigma(W_zx_t+U_zh_{t-1}+b_z)
\end{equation}
where:
\begin{itemize}
    \item $ x_t $ represents the input at time step t.
    \item $ h_{t-1} $ is the hidden state from the previous time step.
    \item $ W_z $  and $ U_z $ are weight matrices.
    \item $ b_z $  is the bias term.
    \item $ \sigma(\cdot) $  is the sigmoid activation function
\end{itemize}
Another key component of GRU is the reset gate, which controls how much of the past hidden state should be forgotten before computing the new candidate hidden state.

The reset gate is defined as:
\begin{equation}
    r_t = \sigma(W_rx_t+U_rh_{t-1}+b_r)
\end{equation}

Here, $ r_t $ determines how much of the past hidden state contributes to the current computation.

Once the reset gate has filtered past information, a candidate hidden state $ \tilde{h}_t $ is computed using:
\begin{equation}
    \tilde{h}_t = \tanh(W_hx_t+U_h(r_t\odot h_{t-1})+b_h)
\end{equation}

The final hidden state $ h_t $ is a weighted combination of the old hidden state $ h_{t-1} $ and the candidate hidden state $ \tilde{h}_t $, modulated by the update gate:
\begin{equation}
    h_t = z_t \odot h_{t-1} + (1-z_t) \odot \tilde{h}_t
\end{equation}

$ z_t $ here determine weather updates with new information or retain past information. 

In the feature extraction layer, the GRU structure consists of two layers, similar to the LSTM design: the first GRU layer with 128 processing units and the second with 64 units. The hidden state from the first GRU layer is passed through batch normalization, while the second GRU layer delivers the final feature vector to the KAN layer.\\

\textbf{Nonlinear Modeling Stage:}\\

The key idea behind Kolmogorov-Arnold Networks (KAN) is that any multi-variable function $ f(x_1, ..., x_n) $ can be transformed into a combination of multiple single-variable functions. By leveraging non-linear transformation functions, KAN is capable of capturing intricate non-linear interactions between different variables, which traditional linear models may struggle to represent effectively.

In this model, LSTM or GRU serves as the feature extraction layer, producing a hidden state vector that encodes temporal dependencies from sequential data. However, the relationships between these hidden state features are often highly complex and non-linear. For instance, in the context of loan default prediction, factors such as a borrower’s credit rating, income level, and debt status do not exhibit a simple linear correlation with the likelihood of default. Instead, these factors interact in a more intricate manner, requiring a model that can effectively decompose and understand their underlying dependencies.

KAN addresses this challenge by decomposing the complex relationships embedded in the hidden state vector. KAN layer follows sequence processing, specifically targeting the output of the second GRU and LSTM layers. Unlike MLPs, which employ fixed node activation functions, KAN utilizes edge-based learnable activation functions \citep{liu2024}. Specifically, each weight in MLP is replaced by a univariate function parameterized as a spline function, thus eliminating the need for linear weights in the network. In particular, a conventional ``input-hidden-output'' structure of an MLP can be substituted by the composition of two KAN layers, satisfying the following equation:

\begin{equation}
    KAN_{output} = \sum^{2n+1}_{q=1}\Phi_q \left(\sum^{n}_{p=1}\phi_{p,q}(x_p)\right)
\end{equation}

where
\begin{itemize}
    \item $ x_p $ represents the p-th element of the hidden state vector.
    \item $\phi_{q,p}: [0,1] \rightarrow \mathbb{R}$ is a learned non-linear function that transforms each individual feature separately.
    \item $\Phi_q:\mathbb{R} \rightarrow \mathbb{R}$ is another non-linear activation function, applied to the aggregated sum to introduce additional flexibility in modeling complex interactions.
\end{itemize}

When extending a single KAN layer into L stacked layers, the model can be represented in a more compact form as:
\begin{equation}
    KAN_{output} = (\Phi_L \circ \Phi_{L-1} \circ ... \circ \Phi_1)x
\end{equation}

In this hierarchical KAN network, each individual KAN layer functions as a small sub-network, progressively transforming the input through multiple levels of non-linear feature extraction. By stacking multiple KAN layers, the model gains enhanced expressive power, enabling it to capture increasingly complex relationships and deeper hierarchical representations within the data.

After the KAN layer, the fully connected layer performs a nonlinear transformation on the univariate functions output by the KAN layer. By applying linear transformations and activation functions, it further processes and refines the extracted features, enabling the model to learn higher-level feature representations.

To enhance generalization and prevent overfitting, a regularization layer, specifically a Dropout layer, is introduced to randomly deactivates a portion of neurons during training. This prevents the model from over-relying on specific neurons, reducing the risk of overfitting, where the model performs well on the training set but struggles with new or unseen data.

The output layer will transform the output of the fully connected layer into a probability value ranging from 0 to 1, representing the predicted probability of default.

\subsection{Dataset and Data Preprocessing}
\subsubsection{Dataset}

The experiment utilizes the Single-Family Loan-Level dataset from Freddie Mac \citep{FreddieMac}. This dataset includes comprehensive details on loan dates, overdue status, and vital information for fraud detection. The target variable, known as default, is binary, based on the current loan delinquency status (CLDS), indicating days overdue since the last installment was due. A CLDS value of 3 or more results in a default value of 1, otherwise 0. This default definition follows the Basel II industry standard \citep{BCBS2006}. Six features per transaction are selected as input variables, detailed in Table~\ref{tab:feature}.

\begin{table}[!ht]
\centering
\caption{Overview of feature descriptions in the Single-Family Loan-Level dataset by Freddie Mac \citep{FreddieMac}}
\begin{tabular}{lp{9cm}}
\toprule
Feature & Description \\ \midrule
\multicolumn{2}{l}{\textbf{Target variable:}} \\
Default & Equals 1 when CLDS is 3 or more, and 0 otherwise.\\
\midrule
\multicolumn{2}{l}{\textbf{Input variables:}} \\
Assistance status code & Type of support arrangement for short-term loan payment relief. \\
Current actual UPB & Indicates the stated final unpaid principal balance (UPB) of the mortgage.\\
Current deferred UPB & The present non-interest accruing UPB of the adjusted loan.\\
Current interest rate & Displays the present interest rate on the mortgage note, including any modifications. \\
Estimated loan to value & Present LTV ratio as determined by Freddie Mac’s AVM value. \\
Interest bearing UPB & The present UPB of the adjusted loan that accumulates interest. \\
\bottomrule
\end{tabular}
\label{tab:feature}
\end{table}


The study uses data from the first quarter of 2019 for training and the first quarter of 2020 for testing. This setup mimics real-world situations where predictions are made on future events using past data, despite potential shifts in data distribution over time. It assesses the model's adaptability to changes, ensuring efficacy in practical cases. Table~\ref{tab:summary} outlines the summary statistics: the training set includes 281,050 entries with an average loan length of 29.11 months and 1.33\% default rate; the test set has 517,778 entries, an average loan length of 30.52 months, and a default rate of 0.81\%.

\begin{table}[!ht]
\centering
\captionsetup{justification=centering}
\caption{Statistics of training set and test set extracted from the Single-Family Loan-Level dataset by Freddie Mac \citep{FreddieMac}}
\begin{tabular}{lcccccc}
\toprule
\multirow{2}*{Sample} & \multirow{2}*{Year} & Sample & No. of & Average Loan & Default & No. of  \\
&& Size & Loans & Length (Months) & Rate & Default\\ \midrule
Training set & 2019 & 8,180,151 & 281,050 & 29.106 & 1.333\% & 3,746\\
Test set & 2020 & 15,804,438 & 517,778 & 30.524 & 0.811\% &4,199 \\ 
\bottomrule
\end{tabular}
\label{tab:summary}
\end{table}

\subsubsection{Data Preprocessing}\label{sec:3.2.2}

The dataset is structured by loan sequence number, with data sorted by the remaining months to legal maturity to satisfy the window size and prediction period requirements. The window length is defined on the basis of the test condition, and sequences shorter than this length are removed. The borrower assistance status code is transformed using One-Hot encoding to convert the categorical variable into a binary vector suitable for model processing. A new feature, known as interest bearing UPB-delta, is derived by calculating the first-order difference of the interest bearing UPB, highlighting the changes between consecutive points. This feature replaces the interest bearing UPB to reduce long-term trends and seasonal effects, employing first-order differencing to highlight short-term variations, following the recommendation by \citet{feature}, as first-order differencing is a common technique in feature
engineering. 

To address class imbalance, random undersampling is applied to training and test datasets, ensuring equal default and non-default instances.  As the focus on resampling is not central to this study, random undersampling was chosen as it has been empirically proven to be the simplest and one of the most competitive resampling methods among resampling techniques \citep{PraveenMahesh2022DetectionOF}. The undersampling procedure retained all the default data, so that sample differences in trials from 20 repetitions were obtained by randomly selecting equal non-default samples.


\subsection{Evaluation Metrics} \label{sec:3.3}

The performance of the model is assessed through standard metrics such as accuracy\citep{ref54}, precision\citep{ref54}, recall \citep{ref54}, F1 score \citep{ref4}, and AUC \citep{ref4}, offering a thorough evaluation of classification effectiveness. These metrics are defined as follows:
\[
\text{Accuracy} = \frac{\text{TP} + \text{TN}}{\text{TP} + \text{TN} + \text{FP} + \text{FN}}
\]

\[
\text{Precision} = \frac{\text{TP}}{\text{TP} + \text{FP}}
\]

\[
\text{Recall} = \frac{\text{TP}}{\text{TP} + \text{FN}}
\]

\[
\text{F1} = 2 \cdot \frac{\text{Precision} \cdot \text{Recall}}{\text{Precision} + \text{Recall}}
\]
\vspace{0.5em}

Accuracy quantifies the ratio of correctly predicted instances among all instances, providing a basic evaluation of the performance of the model. However, in imbalanced datasets, accuracy can be misleading as it often highlights the dominance of the majority class over true predictive capability. Precision denotes the fraction of true positive predictions among all predicted positives, whereas recall indicates the fraction of true positives correctly identified out of all actual positives. The studies by \citet{Cekic_2023} and \citet{Hilal_2021} indicated that the recall rate is frequently emphasized over precision in anomaly detection. In loan default prediction, a high recall rate is vital to ensure that most actual defaulted loans are identified, helping financial institutions manage risk effectively and minimize losses. 

Ideally, a robust model achieves high precision and recall, but there is often a trade-off between them in practice. High precision suggests a cautious model that predicts positive only with high confidence, potentially missing true positives, and thus reducing recall. However, high recall might result from predicting more positives to capture as many true positives as possible, leading to more false positives and reducing precision. This trade-off highlights the necessity to balance these metrics, where focusing on one metric alone might not produce an ideal model. F1 score (harmonic mean of precision and recall) provide a more comprehensive assessment. The classification threshold of F1 is set at 0.5, indicating a balanced class distribution. AUC (Area Under the Curve) assesses the class differentiation ability of a model by summarizing the trade-off between true and false positive rates at various thresholds. An AUC near 1 shows better discrimination. These metrics ensure that both positive and negative classes are addressed, especially in contexts where class trade-offs are critical.

\section{Experiment Design and Results Analysis}

Three experiments were designed to evaluate the performance of the proposed models in three different aspects, including the impact of different feature window lengths (see Section \ref{sec:4.3}), the early prediction capability for default events (see Section \ref{sec:4.2}), and with limited training samples (see Section \ref{sec:4.1}). The evaluation involved comparing the proposed model against the baselines of GRU, LSTM, LSTM-Attention, and LSTM-Transformer. Section \ref{sec:4.4} further discusses model performance across cohorts from different years when using the optimal feature window, data volume, and time interval derived under the three experiments.


\subsection{Feature Window Lengths Performance Analysis} \label{sec:4.3}

To assess how different feature window lengths affect the proposed model, the lengths of \( x \) are set to 12, 15, 18, 21, 24, and 27 months. These windows serve as inputs to predict defaults within the next 3 months, with this 3-month observation period acting as the target variable \( y \). For each window length, the models are trained using historical data from the months specified prior to the observation period. This analysis reveals the trade-off between window length and model performance, highlighting the optimal feature selection configuration in predictive tasks, helping us to understand the extent of the model’s dependence on the feature window length and its ability to adapt on data with different time spans.

Table~\ref{tab:exp3} in the appendix presents the performance comparison of the proposed and baseline models with varying feature window lengths. The data reveals a significant influence of window length on model outcomes on the OOT dataset. In these conditions, GRU-KAN or LSTM-KAN models generally achieve high performance on most metrics. The LSTM-Transformer model demonstrates specific benefits, achieving high precision in shorter windows and higher recall and AUC in longer ones. However, the performance across all metrics exhibits significant fluctuations, highlighting the sensitivity of models to feature window length and potentially indicating challenges in maintaining consistency across diverse input setups.



\begin{figure}[!ht]
\centering
\subfloat[]{\includegraphics[width=0.45\linewidth]{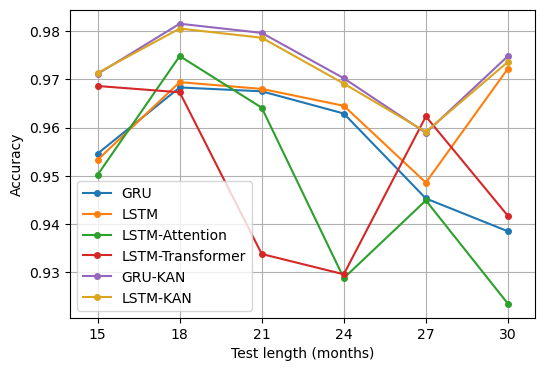} \label{fig:exp3_acc}}
\subfloat[]{\includegraphics[width=0.45\linewidth]{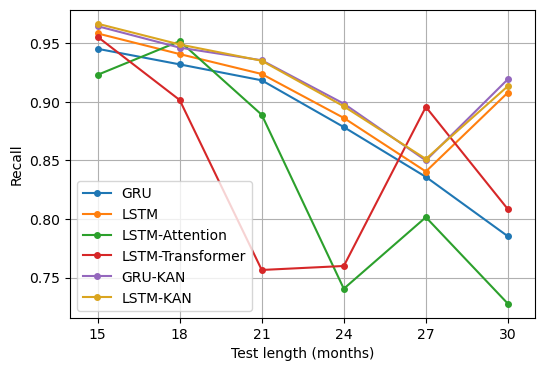} \label{fig:exp3_rec}} \\
\subfloat[]{\includegraphics[width=0.45\linewidth]{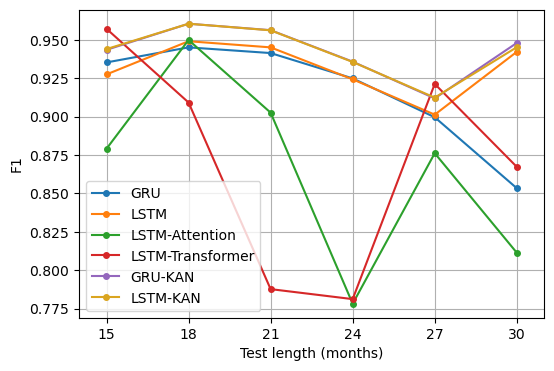} \label{fig:exp3_f1}}
\subfloat[]{\includegraphics[width=0.45\linewidth]{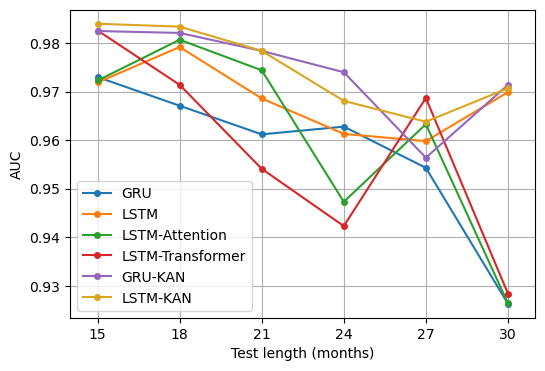} \label{fig:exp3_auc}}
\caption{Line charts depicting (a) accuracy, (b) recall, (c) F1 score, and (d) AUC for each model using varying test window lengths. The horizontal axis indicates the total time series length of the test set, inclusive of the feature window and observation period \( y = 3 \) (3 months).}
\label{fig:exp3}
\end{figure}

Figure \ref{fig:exp3} presents trends in average evaluation metrics from Table \ref{tab:exp3} as the feature window length changes. The horizontal axis shows the total time series length of the test set, which includes both the input feature window length and an observation period \( y = 3 \), which is 3 months for each \( x \) level. The line charts demonstrate the performance variations of different models over various window lengths, emphasizing the feature window length's impact on model performance. In loan default prediction, precision is less important than recall, and its impact is already reflected in the F1 score. Consequently, precision data are excluded from the charts, with detailed results available in the Table~\ref{tab:exp3} in Appendix.

GRU and LSTM show similar trends in accuracy, recall, and F1 scores, except for the longest window length, where LSTM's strong long-term dependency capture makes it more effective with longer time series than GRU. However, integrating GRU with the KAN layer greatly increases its performance by effectively modeling non-linear structures. The performance of GRU-KAN and LSTM-KAN is similar across various metrics, indicating that the KAN layer provides comparable optimization benefits, thus narrowing the feature learning differences of the model. In terms of AUC, the variations between GRU-KAN and LSTM-KAN mirror those between GRU and LSTM, as each performs better depending on the window length, though to different degrees.

Both LSTM-Attention and LSTM-Transformer exhibit significant variability across all metrics as the feature window length increases, indicating reduced robustness. The main difference between these models is the inclusion of a feedforward network in LSTM-Transformer, yet their performance trends are comparable. The difference lies in their window length preferences, with LSTM-Transformer better at managing long time series than LSTM-Attention. Specifically, these models perform best with window lengths of 18 and 27, likely due to optimal alignment with the processing of the attention layer. This situation emphasizes the sensitivity of the models to specific feature window lengths.

Generally, the proposed models consistently showed better performance across different feature window lengths, validating their robustness and adaptability. This improvement in key metrics highlights their strength in capturing complex feature relationships and managing temporal data.

\subsection{Early Prediction Performance Analysis} \label{sec:4.2}

The ``early prediction'' condition is created by introducing a gap between \( x \) and \( y \), where \( x \) represents the window of input characteristics, and \( y \) is the observation period for forecasting. In sections \ref{sec:4.3} and \ref{sec:4.1}, the gap is set to zero, making \( x \) and \( y \) adjacent, which allows the model to use the latest features of the time period for prediction, eliminating any interval between the input and output periods.

\begin{figure}[!htbp]
\centering
\subfloat{%
\resizebox*{10cm}{!}{\includegraphics{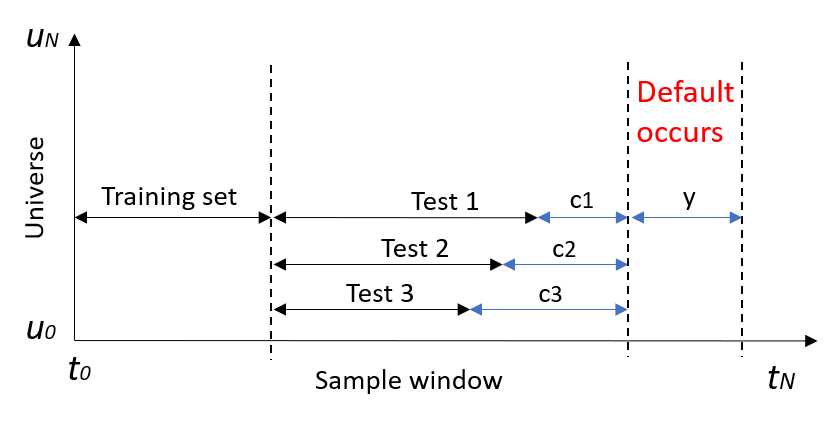}}}
\caption{Overview diagram of experiments using different blank interval configurations for different tests.} \label{fig:task2}
\end{figure}

Blank interval gradients are established at 3, 4, 5, 6, 7, and 8 months to assess the ability of models to forecast defaults by analyzing repayment behavior 3 to 8 months in advance. Both the training and the testing phases utilize the complete dataset. Here, \( y \) is set for 3 months, while the total duration of \( x \) and the interval remain at 21 months, altering the length of the input feature window with the interval. Figure \ref{fig:task2} shows the test set, in which the model predicts defaults for a duration from \( t \) to \( t+3 \) months with data spanning \( t-21 \) to \( t-i \) months, where \( i \) months are denoted as the interval duration.

Table \ref{tab:exp2} provides a systematic comparison of the mean, best value, and standard deviation of five metrics over 20 trials in six models under various prediction intervals. The findings demonstrate that the GRU-KAN model outperformed the LSTM-KAN and other baselines in terms of accuracy, recall, and F1 score at every interval. The GRU-KAN maintained an F1 score advantage due to its high recall, even though its precision was not the highest across all intervals. Regarding AUC, LSTM-KAN performed best only when the blank interval was set to 3 months, with an average value of 0.9278. The GRU-KAN consistently achieved the best results over 4 to 8-month intervals, demonstrating its robustness and excellent performance across varying conditions. Notably, our proposed models achieved the same level of Accuracy, Recall, and F1 at a 5-month blank interval as other baseline models do at a 3-month interval. This indicates that the proposed models can deliver equally accurate default predictions two months earlier. This additional reaction time provides financial institutions with a significant advantage, enabling them to implement early risk management strategies and response measures more effectively. Such a capability is crucial for mitigating potential losses and optimizing decision-making processes.

\begin{figure}[!ht]
\centering
\subfloat[]{\includegraphics[width=0.45\linewidth]{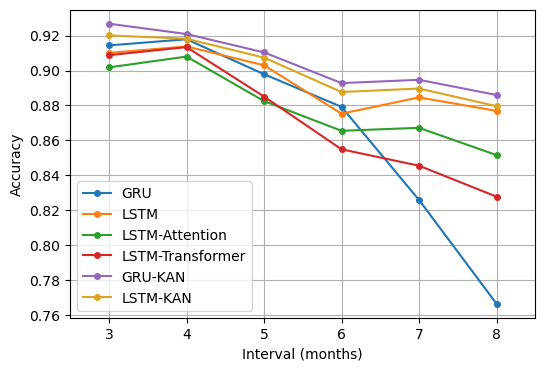} \label{fig:exp2_acc}}
\subfloat[]{\includegraphics[width=0.45\linewidth]{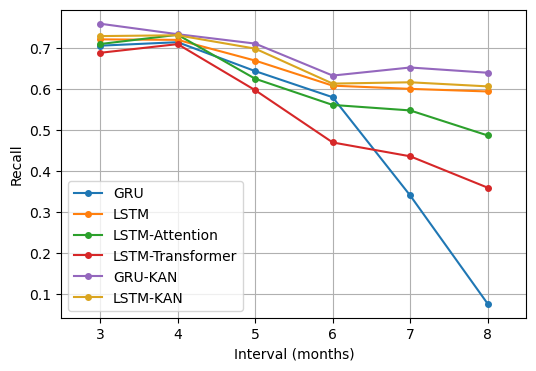} \label{fig:exp2_rec}} \\
\subfloat[]{\includegraphics[width=0.45\linewidth]{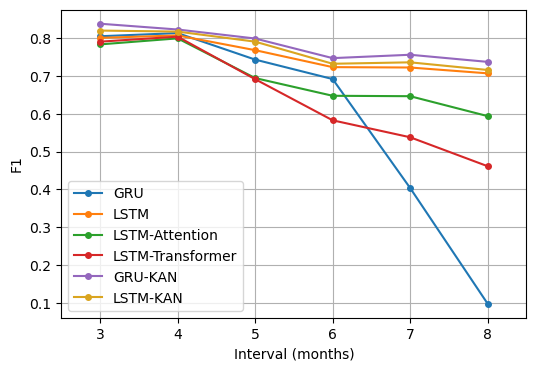} \label{fig:exp2_f1}}
\subfloat[]{\includegraphics[width=0.45\linewidth]{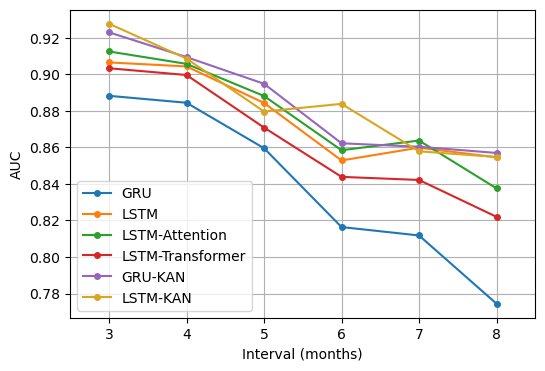} \label{fig:exp2_auc}}
\caption{Line charts displaying (a) accuracy, (b) recall, (c) F1 score, and (d) AUC for each model under different blank intervals.}
\label{fig:exp2}
\end{figure}

Figure \ref{fig:exp2} illustrates the average value from Table \ref{tab:exp2} to demonstrate the declining trend. With longer blank intervals, input information becomes more detached from default events, leading to a marked drop in model performance. Yet, GRU-KAN and LSTM-KAN consistently outperform others across metrics. As for decline rates, when the interval expands from 3 to 8 months, GRU-KAN and LSTM-KAN accuracies fall by about 4.40\%, while GRU, LSTM, LSTM-Attention, and LSTM-Transformer accuracies decrease by 16.17\%, 3.65\%, 5.57\%, and 8.91\%, respectively. Despite LSTM having the smallest decline, its accuracy is still below that of the proposed models.

Similarly, LSTM exhibits minimal decline, whereas GRU shows the largest reduction in AUC. The AUC values of GRU-KAN and LSTM-KAN decrease by 7.16\% and 7.87\%, respectively, during the 3 to 8-month gap. In contrast, the baseline models GRU, LSTM, LSTM-Attention, and LSTM-Transformer experience declines of 12.82\%, 5.74\%, 8.22\%, and 9.01\%. Thus, GRU-KAN and LSTM-KAN display greater robustness and adaptability over different interval settings.

For Recall and F1, GRU-KAN and LSTM-KAN sustained high performance even after some reduction, showing F1 scores above 0.7 and recall values above 0.6. In contrast, baseline models, especially GRU, experienced a significant drop in F1 (-87.80\%) and recall (-89.17\%) as the blank interval increased from 3 to 8 months, nearly losing the prediction ability. Similarly, LSTM-Attention and LSTM-Transformer saw recall fall to about 0.4 at an 8-month interval. However, LSTM showed relatively stable performance, with a decline similar to the proposed models, although its absolute values were slightly lower.

LSTM-Attention and LSTM-Transformer exhibit the lowest accuracy, recall, and F1 scores, suggesting that the weighting of the Attention layer and the self-attention mechanisms of the Transformer perform poorly under ``early prediction'' conditions. The Attention layer might struggle to emphasize crucial temporal details when handling long intervals, leading to a decline
in model efficacy. While the Transformer layer is strong in global modeling, its complexity can add extra computational burden for short sequences or sparse data, thereby diminishing the LSTM's natural sequential modeling capability.

The performance of GRU diminishes with significant input-output sequence gaps, likely due to constraints in its gating mechanism. The update and reset gates of GRU aim to model long-term dependencies, but long intervals can result in the loss of crucial temporal information. In addition, the design of GRUs has difficulty in capturing intricate patterns and monitoring dynamic characteristics over time, leading to a decreased prediction accuracy. Conversely, the KAN layer focuses on capturing non-linear relationships and dynamically learning activation functions to effectively extract features linked to the target event, even with lengthy intervals and sparse data. This method addresses performance challenges resulting from information loss, thereby improving the model's robustness and predictive precision.


The peak metric values of the proposed models are consistently lower than those in Section \ref{sec:4.1} for all configurations of blank intervals, with the default interval set to 0. This suggests that introducing a blank interval between input features and the observation period presents challenges such as feature sparsity or information loss due to the lengthy intervals. Consequently, models are unable to directly utilize features near the prediction target, resulting in reduced performance.

\subsection{Sample Size Performance Analysis}\label{sec:4.1}

Various sample size configurations were employed to evaluate the feasibility and predictive capability of the proposed models under limited data conditions. The total duration of the training and testing window is set at 18 months, with a feature window of 15 months, following the optimal length obtained in Section \ref{sec:4.3}. No blank interval is utilized as early prediction is not included here. Sequences less than 18 months (15+3 months) are excluded due to inadequate historical data for effective training. Longer sequences are truncated at 18 months for consistency. Table \ref{tab:samplesize} presents the count of loans lasting at least 18 months, grouped by loan sequence number, along with the number of defaulted loans and the corresponding default rate, for each sample size of repayment records (500,000, 1,000,000, 1,500,000, 2,000,000, 3,000,000, and 5,000,000 entries).



\begin{table}[h]
\centering
\captionsetup{justification=centering}
\caption{Configurations overview for various sample sizes for training and test sets.}
\begin{tabular*}{\textwidth}{@{\extracolsep\fill}lcccccc}
\toprule%
& \multicolumn{3}{@{}c@{}}{Training set (2019)} & \multicolumn{3}{@{}c@{}}{Test set (2020)} \\\cmidrule{2-4}\cmidrule{5-7}%
& \multicolumn{1}{c}{No. of} & \multicolumn{1}{c}{No. of} & Default & \multicolumn{1}{c}{No. of} & \multicolumn{1}{c}{No. of} & Default \\
Sample size & Eligible  Loans & Default & \multicolumn{1}{c}{Rate} & Eligible  Loans & Default & \multicolumn{1}{c}{Rate} \\ \midrule
500,000 & 10,906 & 249 & 2.283\% & 10,777 & 149 & 1.383\% \\
1,000,000 & 21,902 & 464 & 2.119\% & 21,521 & 300 & 1.394\% \\
1,500,000 & 33,008 & 706 & 2.139\% & 32,292 & 432 & 1.338\% \\
2,000,000 & 44,135 & 934 & 2.116\% & 43,134 & 584 & 1.354\% \\
3,000,000 & 66,427 & 1,373 & 2.067\% & 64,782 & 862 & 1.331\% \\
5,000,000 & 110,946 & 2,271 & 2.047\% & 108,086 & 1,476 & 1.366\% \\
\botrule
\end{tabular*}
\label{tab:samplesize}
\end{table}

Table~\ref{tab:exp1} summarizes the mean and best outcomes of 20 independent trials with standard deviation, analyzing differences in accuracy, precision, recall, F1 score, and AUC between the proposed and baseline models with varying sample sizes. The findings show that performance gaps between models are more noticeable with smaller sample sizes, which diminish as the sample size grows. GRU-KAN and LSTM-KAN demonstrate comparable performance, alternating in superiority across different sample size configurations. Both models consistently outperform the baseline models, displaying improved performance and stability across sample sizes, illustrating their effectiveness with limited data. A comparison with the results presented in Table~\ref{tab:exp2} further demonstrates that including a blank interval generally enables GRU-KAN to outperform LSTM-KAN compared to the absence of such an interval.

\begin{figure}[!ht]
\centering
\subfloat[]{\includegraphics[width=0.45\linewidth]{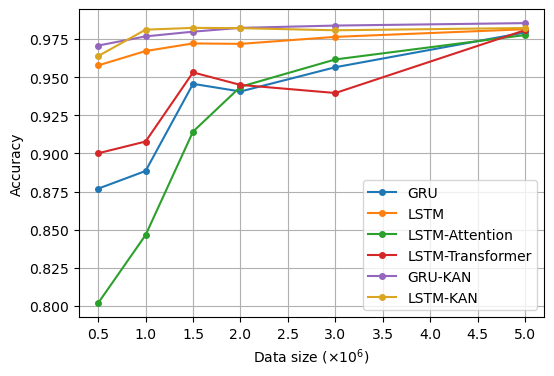} \label{fig:exp1_acc}}
\subfloat[]{\includegraphics[width=0.45\linewidth]{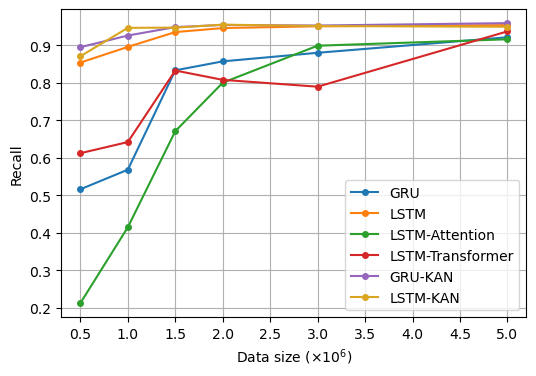} \label{fig:exp1_rec}} \\
\subfloat[]{\includegraphics[width=0.45\linewidth]{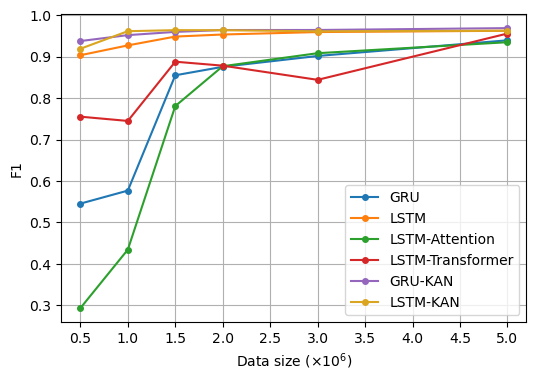} \label{fig:exp1_f1}}
\subfloat[]{\includegraphics[width=0.45\linewidth]{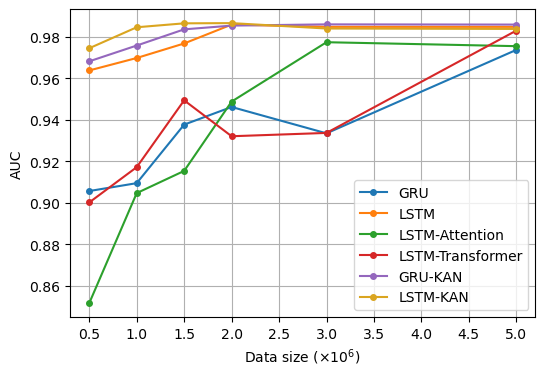} \label{fig:exp1_auc}}
\caption{Line charts displaying (a) accuracy, (b) recall, (c) F1 score, and (d) AUC for each model in different sample sizes.}
\label{fig:exp1}
\end{figure}

Figure~\ref{fig:exp1} displays line charts based on the average results of Table~\ref{tab:exp1}, showing trends between varying sample sizes. LSTM emerges as the highest performing baseline, closely trailing GRU-KAN and LSTM-KAN in all metrics, suggesting that KAN improves GRU more than LSTM. The KAN layer effectively narrows the performance gap between GRU and LSTM in this task. GRU-KAN, LSTM-KAN, and LSTM exhibit stable trends, with low standard deviations across all sample sizes. In contrast, GRU, LSTM-Attention, and LSTM-Transformer perform poorly with small samples, as indicated by lower metric values and higher standard deviations, suggesting that they require larger datasets for reliable performance.


\subsection{Model Generalization Performance Analysis} \label{sec:4.4}

To evaluate the generalization of the proposed models in different data groups, experiments were performed using cohorts from different years, where each cohort contains monthly repayment data for loan applicants starting from its specific year, making each cohort an independent dataset. Six training-test pairs were formed using 2018 data for training and data from 2019 to 2022 for testing, with 2019 data also used for training and 2021 and 2022 data for testing. Data from 2023 onward were excluded, as they were not yet fully updated on the existing dataset.

The optimal sample size and feature window length from the previous results were applied to all training and test sets from different cohorts. To evaluate model performance in the ``early prediction'' context, a blank interval was used. Specifically, the feature window length was set at 15 months, with a 3-month blank interval and a 3-month observation period. Considering that the LSTM model's performance stabilized with 1.5 million records, the first 1.5 million repayment records from each cohort were used for training and testing to assess the model's generalization across cohorts. Table \ref{table4} presents the number of unique loans within the first 1.5 million records for each cohort, the count conforming to the 21-month data length criterion (15 + 3 + 3), the number classified as defaults, and their respective default rates.

\begin{table}[!ht]
\centering
\captionsetup{justification=centering}
\caption{Summary of total loan configurations and the count of loans meeting the 21-month eligibility criterion under different cohorts.}
\begin{tabular}{ccccr}
\toprule
 \multicolumn{1}{c}{Year} of & \multicolumn{1}{c}{No. of} & \multicolumn{1}{c}{No. of} & \multicolumn{1}{c}{No. of} & Default \\
Cohort & Unique Loans & Eligible  Loans & Default & \multicolumn{1}{c}{Rate}  \\ \midrule
2018	&	33,690	&	30,007	&	558	&	1.8596\%	\\
2019	&	51,958	&	28,850	&	698	&	2.4194\%	\\
2020	&	50,511	&	29,339	&	423	&	1.4418\%	\\
2021	&	46,779	&	41,951	&	229	&	0.5459\%	\\
2022	&	67,514	&	63,387	&	580	&	0.9150\%	\\
\bottomrule
\end{tabular}
\label{table4}
\end{table}

\begin{figure}[!ht]
\centering
\subfloat[]{\includegraphics[width=0.45\linewidth]{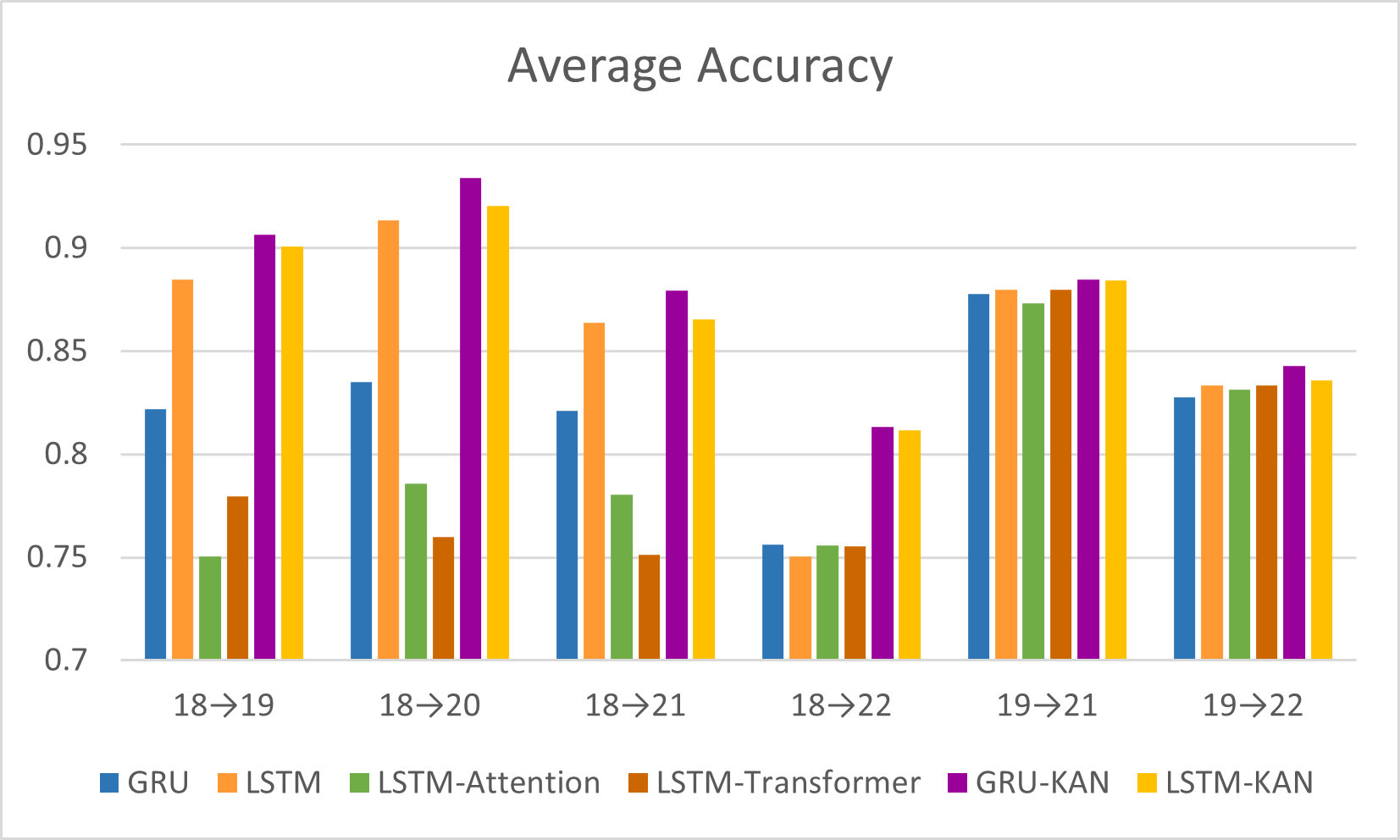} \label{fig:exp4_acc}}
\subfloat[]{\includegraphics[width=0.45\linewidth]{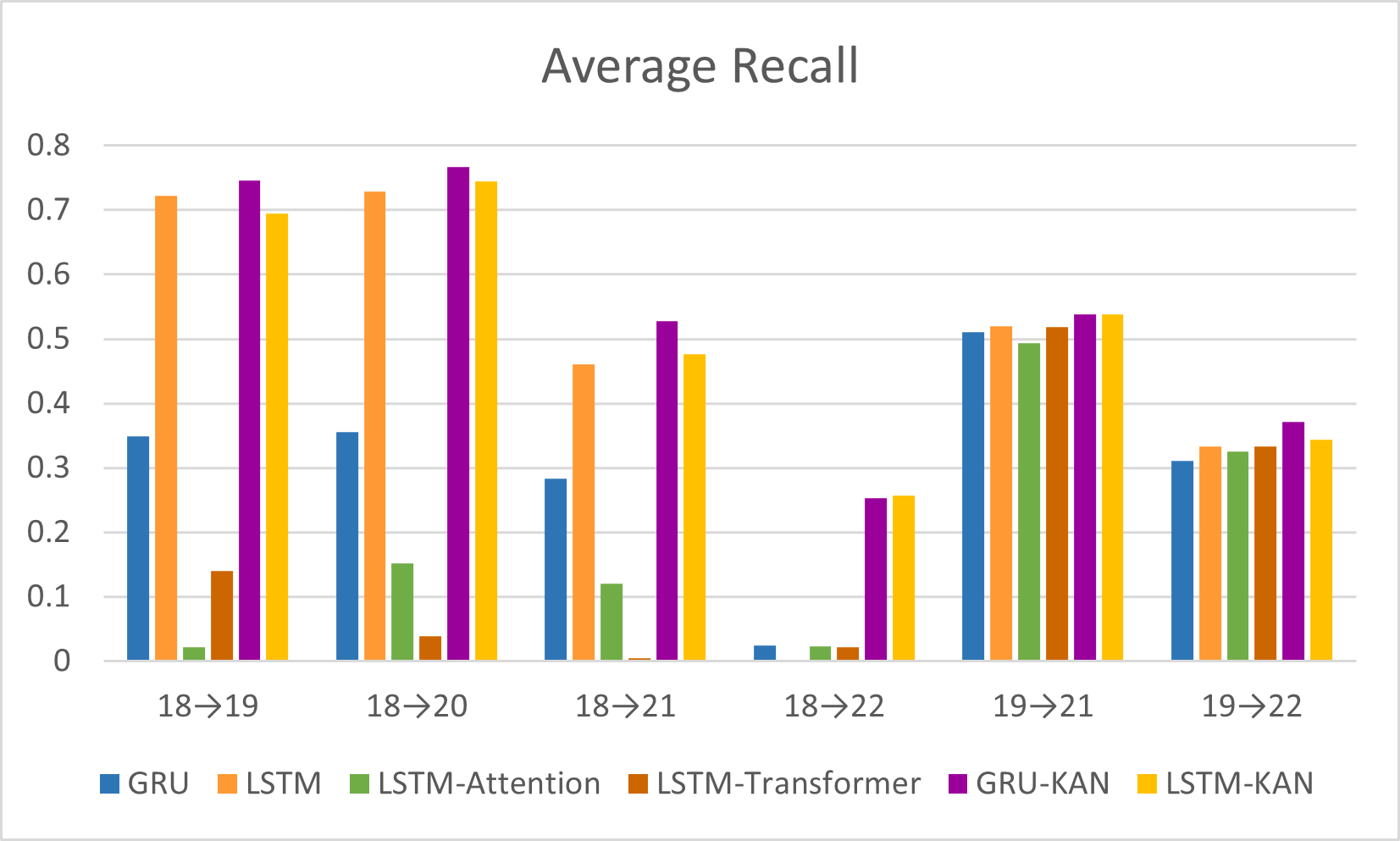} \label{fig:exp4_rec}} \\
\subfloat[]{\includegraphics[width=0.45\linewidth]{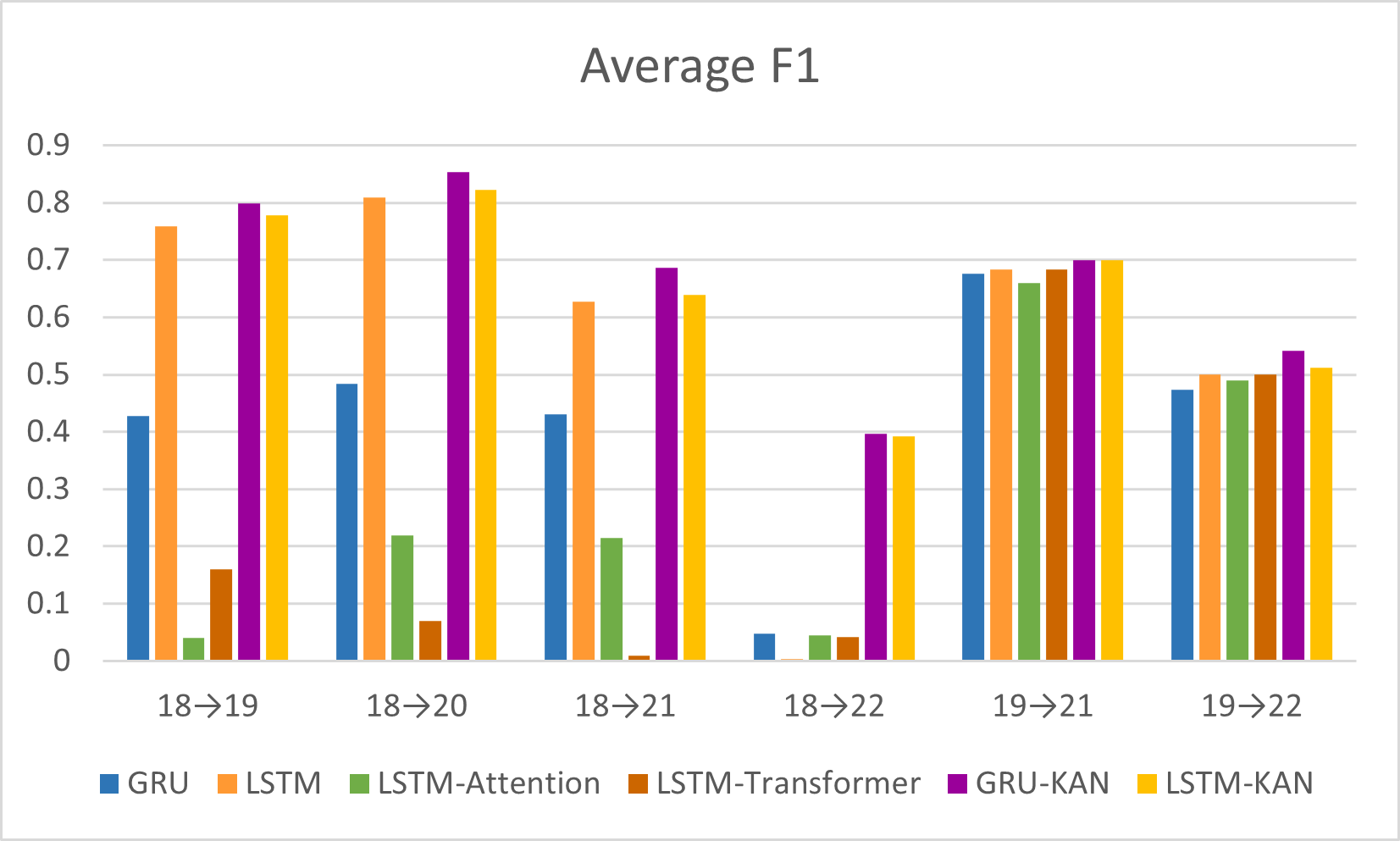} \label{fig:exp4_f1}}
\subfloat[]{\includegraphics[width=0.45\linewidth]{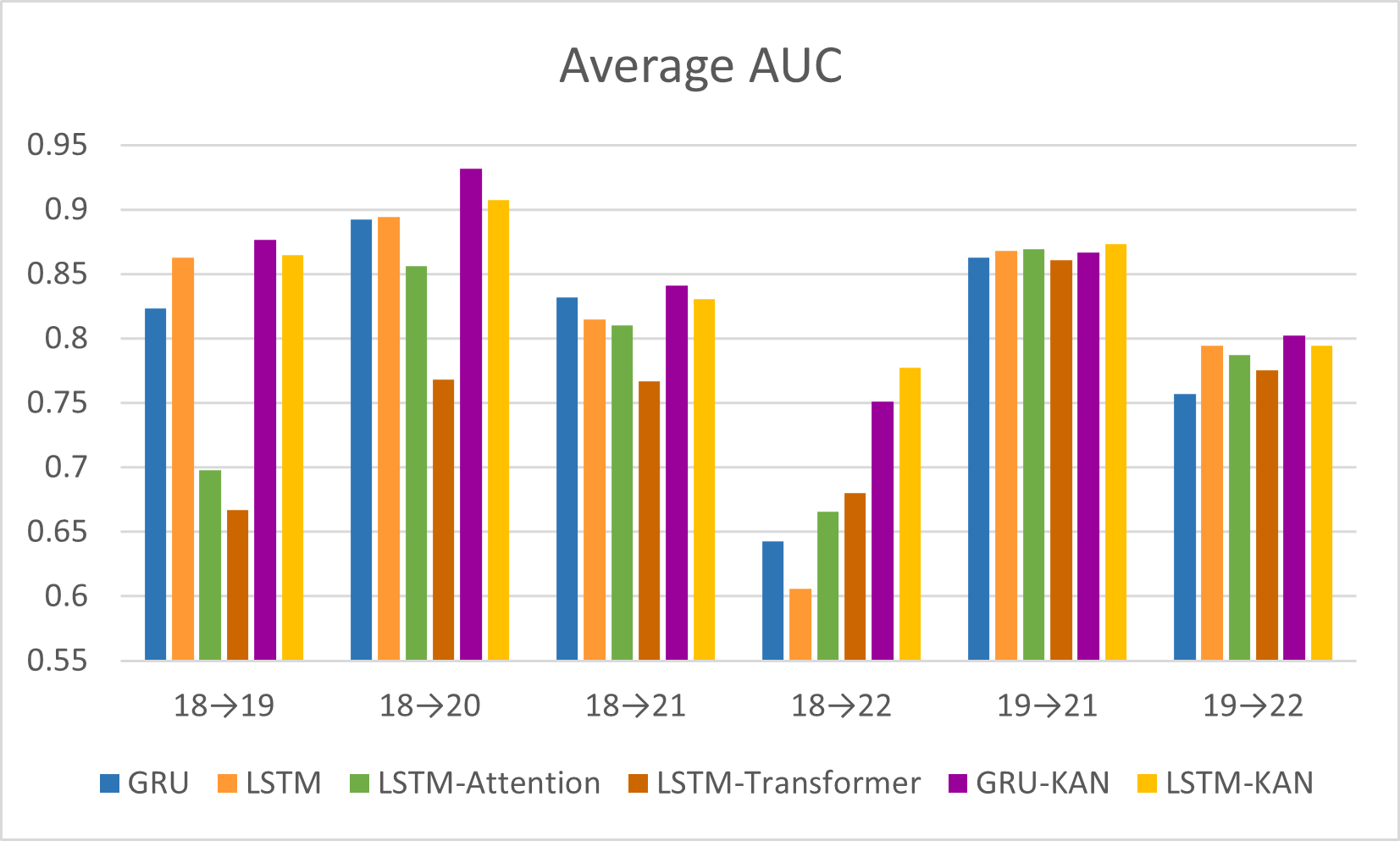} \label{fig:exp4_auc}}
\caption{Bar charts displaying (a) accuracy, (b) recall, (c) F1 score, and (d) AUC for each model across training and test sets from various years. The number preceding the ``$\to$'' signifies the training data year, while the number following it denotes the test data year.}
\label{fig:exp4}
\end{figure}

Table~\ref{tab:exp4} summarizes the average, best, and standard deviation for the six training-test set pairs (six cohort pairs) over 20 independent trials. Figure~\ref{fig:exp4} shows a bar chart derived from the average values in Table~\ref{tab:exp4}. The results indicate that GRU-KAN consistently achieves a higher accuracy and F1 score across all cohort pairs, with the highest recall in five pairs and the best AUC in four, highlighting its robust generalization. In contrast, GRU shows the highest precision in four pairs but struggles with precision-recall balance, resulting in lower overall F1 scores.

An analysis of the impact of the time gap between the training and test sets on performance trends reveals a significant drop in all metrics when predicting 2022 loan defaults using 2018 data, compared to predictions made using 2019 data. This suggests a concept drift \citep{2023_48}, with data distributions evolving over time, leading to new patterns in the future versus historical data. However, the decline in performance with increased temporal gaps is not entirely linear, indicating that the impact of concept drift on model performance does not strictly correlate with the temporal gap. In the evaluation of four cohort pairs using 2018 data, the proposed models consistently outperformed the baseline, with GRU-KAN showing the best performance. This indicates that the proposed models retain strong adaptability and generalization in cross-year predictions, despite the concept drift. For the pair using data from 2019 as the training set, model performance differences decreased, as shown in Sections \ref{sec:4.3} to \ref{sec:4.2}, where the 2019-2020 cohort showed smaller performance gaps than the 2018 training scenario. However, GRU-KAN consistently surpassed other models in the majority of metrics, demonstrating its high generalization under different conditions.

\section{Limitations and Future Work}

This study used only the Freddie Mac dataset. Despite its large size and the representation of real-world loan data, future work should test the models with additional datasets to evaluate their robustness and applicability in varying financial contexts. The study addressed class imbalance through undersampling. Future work could explore a wider array of resampling methods, such as synthetic data generation with SMOTE or GANs, to improve imbalance handling and model performance. Furthermore, future work might focus on developing adaptive models that can effectively address concept drifts, allowing them to adjust dynamically to changing data distributions for more reliable and accurate predictions in environments with substantial temporal shifts.

\section{Conclusion}

The GRU-KAN and LSTM-KAN models improve proactive risk management for loan default prediction by allowing reliable default prediction months in advance and effectively managing OOT data. This study also fills a gap by applying LSTM-Transformer as a baseline to examine Transformer models to predict loan default. The study indicates that these models outperform the baseline models in various metrics, with GRU-KAN consistently demonstrating greater performance and robustness, independent of feature window lengths, small training datasets, and early prediction conditions. These attributes are essential in financial environments where prompt and precise predictions are critical for minimizing risks and making informed decisions. KAN-based models significantly improve existing methods by effectively adapting to concept drift, resulting in greater predictive accuracy and efficiency when analyzing recent test data with prior training data. This advancement makes them more practical and reliable for complex financial applications.



\section{List of abbreviations}
\begin{table}[h!]
\centering
\renewcommand{\arraystretch}{1.5} 
\begin{tabular*}{0.7\textwidth}{@{\extracolsep\fill}ll}
\toprule
\textbf{Abbreviation} & \textbf{Definition} \\ 
\midrule
AUC & area under the ROC curve \\
CLDS & current loan delinquency status \\
CNN & convolutional neural networks \\
GRU  & gated recurrent units \\
KAN  & Kolmogorov-Arnold networks \\
LSTM  & long-short-term memory \\
MLP & multi-layer perceptron \\
OOS & out-of-sample \\
OOT & out-of-time \\
UPB & unpaid principal balance \\
\bottomrule
\end{tabular*}
\end{table}

\newpage
\section*{Notes on Contributors}
\textbf{Yue Yang}: Conceptualization, Data curation, Formal analysis, Methodology, Software, Visualization, Writing – original draft, Writing – review and editing; \textbf{Zihan Su}: Methodology, Software, Validation, Writing – original draft; \textbf{Ying Zhang}: Methodology, Software, Validation, Writing – original draft; \textbf{Chang Chuan Goh}: Software, Visualization; \textbf{Yuxiang Lin}: , Writing – original draft, Validation; \textbf{Boon Giin Lee}: Project administration, Resources, Supervision, Writing – review and editing, Funding acquisition; \textbf{Anthony Graham Bellotti}: Project administration, Resources, Supervision, Writing – review and editing, Funding acquisition. 

\section*{Declaration of Competing Interest}

The authors declare that they have no known competing financial interests or personal relationships that could have appeared to influence the work reported in this paper.

\section*{Data availability}

Availability of data and material: The dataset analyzed during the current study are available in the Freddie Mac repository, \url{https://www.freddiemac.com/research/datasets/sf-loanlevel-dataset}.

\section*{Acknowledgment}

his work was supported by the Ningbo Science and Technology (S\&T) Bureau under Grant 2021B-008-C.

\newpage
\bibliography{main}

\begin{appendices}

\section*{Appendix}\label{secA1}
\setcounter{table}{0}
\renewcommand{\thetable}{A\arabic{table}}
\setcounter{figure}{0}
\renewcommand{\thefigure}{A\arabic{figure}}

\begin{sidewaystable}[!ht]
\tiny
\centering
\caption{Results of model performance based on 20 trials using different input feature window lengths}
\begin{tabular}{clccccc}
\toprule
Test length & Model & Accuracy	(Best, Std) & Precision (Best, Std)	& Recall (Best, Std) & F1	(Best, Std) & AUC (Best, Std) \\ \midrule
15	&	GRU	&	0.9546 (0.9594, 0.0021)	&	0.9256 (0.9479, 0.0118)	&	0.9453 (0.9678, 0.0074)	&	0.9353 (0.9426, 0.0034)	&	0.9730 (0.9769, 0.0031)	\\
&	LSTM	&	0.9533 (0.9587, 0.0026)	&	0.8985 (0.9174, 0.0099)	&	0.9584 (0.9614, 0.0017)	&	0.9275 (0.9309, 0.0047)	&	0.9719 (0.9834, 0.0009)	\\
&	LSTM-Attention	&	0.9502 (0.9679, 0.0472)	&	0.8391 (0.9121, 0.1979)	&	0.9231 (0.9751, 0.2173)	&	0.8791 (0.9377, 0.2070)	&	0.9723 (0.9803, 0.0267)	\\
&	LSTM-Transformer	&	0.9686 (0.9721, 0.0024)	&	\textbf{0.9489} (0.9545, 0.0044)	&	0.9358 (0.9473, 0.0090)	&	0.9423 (0.9539, 0.0042)	&	0.9725 (0.9816, 0.0014)	\\
&	\textbf{GRU-KAN}	&	0.9711 (0.9764, 0.0039) 	&	0.9235 (0.9430, 0.0147) 	&	0.9646 (0.9682, 0.0036) 	&	0.9435 (0.9534, 0.0071) 	&	0.9825 (0.9854, 0.0039) 	\\
&	\textbf{LSTM-KAN} &	\textbf{0.9713} (0.9731, 0.0012) & 0.9224 (0.9283, 0.0048) & \textbf{0.9667} (0.9700, 0.0017) & \textbf{0.9440} (0.9473, 0.0023) 	&	\textbf{0.9840} (0.9843, 0.0003) 	\\[1em]
18	&	GRU	&	0.9683 (0.9742, 0.0010)	&	0.9586 (0.9626, 0.0049)	&	0.9320 (0.9412, 0.0069)	&	0.9451 (0.9565, 0.0061)	&	0.9671 (0.9743, 0.0082)	\\
&	LSTM	&	0.9694 (0.9805, 0.0014)	&	0.9575 (0.9696, 0.0073)	&	0.9408 (0.9544, 0.0036)	&	0.9491 (0.9519, 0.0025)	&	0.9792 (0.9842, 0.0019)	\\
&	LSTM-Attention	&	0.9748 (0.9797, 0.0048)	&	0.9482 (0.9682, 0.0202)	&	\textbf{0.9518} (0.9587, 0.0041)	&	0.9499 (0.9590, 0.0089)	&	0.9807 (0.9830, 0.0017)	\\
&	LSTM-Transformer	&	0.9673 (0.9806, 0.0512)	&	0.9169 (0.9724, 0.2159)	&	0.9016 (0.9560, 0.2122)	&	0.9092 (0.9608, 0.2140)	&	0.9714 (0.9837, 0.0489)	\\
&	\textbf{GRU-KAN}	&	\textbf{0.9815} (0.9808, 0.0002) & \textbf{0.9751} (0.9808, 0.0031) 	&	0.9463 (0.9507, 0.0035) 	&	\textbf{0.9606} (0.9611, 0.0004) 	&	0.9821 (0.9839, 0.0013) 	\\
&	\textbf{LSTM-KAN}	&	0.9805 (0.9807, 0.0001) 	&	0.9724 (0.9738, 0.0008) 	&	0.9490 (0.9503, 0.001) &	0.9605 (0.9609, 0.0002) &	\textbf{0.9834} (0.9853, 0.0014) 	\\[1em]
21	&	GRU	&	0.9675 (0.9693, 0.0014)	&	0.9655 (0.9709, 0.0019)	&	0.9184 (0.9259, 0.0057)	&	0.9414 (0.9456, 0.0029)	&	0.9612 (0.9693, 0.0052)	\\
&	LSTM	&	0.9680 (0.9699, 0.0008)	&	0.9683 (0.9776, 0.0063)	&	0.9238 (0.9443, 0.0052)	&	0.9451 (0.9478, 0.0015)	&	0.9686 (0.9713, 0.0038)	\\
&	LSTM-Attention	&	0.9641 (0.9771, 0.0504)	&	0.9166 (0.9745, 0.2159)	&	0.8890 (0.9438, 0.2093)	&	0.9026 (0.9534, 0.2125)	&	0.9744 (0.9819, 0.0241)	\\
&	LSTM-Transformer	&	0.9338 (0.9804, 0.0822)	&	0.8264 (0.9922, 0.3563)	&	0.7566 (0.9460, 0.3378)	&	0.7877 (0.9600, 0.3437)	&	0.9541 (0.9832, 0.0677)	\\
&	\textbf{GRU-KAN} & \textbf{0.9796} (0.9793, 0.0003)  &	0.9780 (0.9807, 0.002) & \textbf{0.9355} (0.9379, 0.0021) & \textbf{0.9563} (0.9576, 0.0007) &	\textbf{0.9785} (0.9843, 0.0062) \\
&	\textbf{LSTM-KAN}	&	0.9786 (0.9789, 0.0002)  & \textbf{0.9783} (0.9834, 0.0034) 	&	0.9351 (0.9400, 0.0027) 	&	0.9562 (0.9568, 0.0004) 	&	0.9784 (0.9824, 0.0031) 	\\[1em]
24	&	GRU	&	0.9629 (0.9676, 0.0036)	&	0.9765 (0.985, 0.0059) 	&	0.8785 (0.8945, 0.0159)	&	0.9249 (0.9360, 0.0081)	&	0.9628 (0.9736, 0.0064)	\\
&	LSTM	&	0.9645 (0.9702, 0.0039)	&	0.9657 (0.9797, 0.0152)	&	0.8863 (0.8934, 0.0055) 	&	0.9243 (0.9102, 0.0047)	&	0.9613 (0.9664, 0.0033)	\\
&	LSTM-Attention	&	0.9288 (0.9677, 0.0784)	&	0.8712 (1.0000, 0.2983)	&	0.7406 (0.9061, 0.3235)	&	0.7782 (0.9334, 0.3372)	&	0.9473 (0.9750, 0.0396)	\\
&	LSTM-Transformer	&	0.9296 (0.9720, 0.0797)	&	0.8052 (0.9771, 0.3479)	&	0.7600 (0.9335, 0.3339)	&	0.7812 (0.9433, 0.3396)	&	0.9423 (0.9737, 0.0579)	\\
&	\textbf{GRU-KAN} & \textbf{0.9702} (0.9699, 0.0004) & \textbf{0.9833} (0.9880, 0.0031)	&	0.8985 (0.9087, 0.0064) 	&	0.9358 (0.9376, 0.0011) & \textbf{0.9741} (0.9801, 0.0048) \\
&	\textbf{LSTM-KAN} &	0.9691 (0.9701, 0.0005) 	&	0.9787 (0.9875, 0.0052) & \textbf{0.9025} (0.9123, 0.0053) & \textbf{0.9391} (0.9458, 0.0042) 	&	0.9712 (0.9774, 0.0045) 	\\[1em]
27	&	GRU	&	0.9453 (0.9493, 0.0027)	&	0.9736 (0.9871, 0.0025)	&	0.8360 (0.8547, 0.0121)	&	0.8996 (0.9129, 0.0065)	&	0.9543 (0.9587, 0.0040)	\\
&	LSTM	&	0.9486 (0.9402, 0.0006)	&	0.9714 (0.9898, 0.0075)	&	0.8406 (0.8558, 0.0070)	&	0.9013 (0.9103, 0.0015)	&	0.9598 (0.9738, 0.0136)	\\
&	LSTM-Attention	&	0.9449 (0.9597, 0.0223)	&	0.9730 (0.9885, 0.0112)	&	0.8016 (0.8619, 0.0886)	&	0.8762 (0.9139, 0.0667)	&	0.9633 (0.9751, 0.0185)	\\
&	LSTM-Transformer &	\textbf{0.9623} (0.9717, 0.0143) & 0.9509 (0.9682, 0.0166) & \textbf{0.8957} (0.9319, 0.0604) & \textbf{0.9213} (0.9423, 0.0346)	& \textbf{0.9687} (0.9743, 0.0122) \\
&	\textbf{GRU-KAN}	&	0.9589 (0.9600, 0.0007) 	&	\textbf{0.9836} (0.9881, 0.0074) 	&	0.8500 (0.8619, 0.0055) 	&	0.9119 (0.9145, 0.0013) 	&	0.9564 (0.9734, 0.0225) 	\\
&	\textbf{LSTM-KAN}	&	0.9591 (0.9603, 0.0005) 	&	0.9829 (0.9881, 0.0033) 	&	0.8513 (0.8589, 0.0043) 	&	0.9124 (0.9154, 0.0013) 	&	0.9638 (0.9744, 0.0102) 	\\[1em]
30	&	GRU	&	0.9385 (0.9763, 0.0461)	&	0.9584 (0.9820, 0.0356)	&	0.7853 (0.9221, 0.1741)	&	0.8534 (0.9511, 0.1247)	&	0.9263 (0.9708, 0.0655)	\\
	&	LSTM	&	0.9722 (0.9758, 0.0021)	&	0.9793 (0.9915, 0.0076)	&	0.9080 (0.9257, 0.0110)	&	0.9422 (0.9499, 0.0046)	&	0.9699 (0.9726, 0.0019)	\\
	&	LSTM-Attention	&	0.9235 (0.9755, 0.0497)	&	0.9530 (0.9806, 0.0290)	&	0.7278 (0.9284, 0.1946)	&	0.8114 (0.9498, 0.1383)	&	0.9264 (0.9707, 0.0408)	\\
	&	LSTM-Transformer	&	0.9417 (0.9726, 0.0330)	&	0.9512 (0.9787, 0.0238)	&	0.8084 (0.9233, 0.1339)	&	0.8673 (0.9437, 0.0913)	&	0.9284 (0.9689, 0.0436)	\\
	&	\textbf{GRU-KAN}	&	\textbf{0.9748} (0.9764, 0.0011)	&	0.9785 (0.9880, 0.0056)	&	\textbf{0.9194} (0.9292, 0.0079)	&	\textbf{0.9480} (0.9515, 0.0026)	&	\textbf{0.9714} (0.9753, 0.0023)	\\
	&	\textbf{LSTM-KAN} &	0.9736 (0.9774, 0.0032)	&	\textbf{0.9795} (0.9895, 0.0069)	&	0.9135 (0.9308, 0.0153)	&	0.9452 (0.9536, 0.0072)	&	0.9707 (0.9734, 0.0014)		\\
\bottomrule
\end{tabular}
\label{tab:exp3}
\end{sidewaystable}

\begin{sidewaystable}[!ht]
\tiny
\centering
\caption{Results of model performance based on 20 trials using different blank interval.}
\begin{tabular}{clccccc}
\toprule
Interval & Model & Accuracy	(Best, Std) & Precision (Best, Std)	& Recall (Best, Std) & F1	(Best, Std) & AUC (Best, Std) \\ \midrule
3	&	GRU	&	0.9144 (0.9191, 0.0039)	&	0.9353 (0.9432, 0.0157)	&	0.7066 (0.7258, 0.0098)	&	0.8049 (0.8177, 0.0084)	&	0.8883 (0.9174, 0.0241)	\\
&	LSTM	&	0.9101 (0.9296, 0.0065)	&	0.9006 (0.9509, 0.0350)	&	0.7220 (0.7892, 0.0236)	&	0.8007 (0.8486, 0.0130)	&	0.9066 (0.9357, 0.0317)	\\
&	LSTM-Attention	&	0.9018 (0.9094, 0.0028)	&	0.8731 (0.8981, 0.0124)	&	0.7106 (0.7237, 0.0107)	&	0.7834 (0.7987, 0.0063)	&	0.9126 (0.9278, 0.0188)	\\
&	LSTM-Transformer	&	0.9088 (0.9165, 0.0050)	&	0.9295 (0.9482, 0.0312)	&	0.6892 (0.7575, 0.0288)	&	0.7906 (0.8109, 0.0133)	&	0.9034 (0.9288, 0.0267)	\\
&	\textbf{GRU-KAN}	&	\textbf{0.9268} (0.9440, 0.0090) 	&	0.9351 (0.9421, 0.0070) 	&	\textbf{0.7602} (0.8433, 0.0426) 	&	\textbf{0.8380} (0.8828, 0.0236) &	0.9231 (0.9303, 0.0085) \\
&	\textbf{LSTM-KAN}	&	0.9201 (0.9320, 0.0047)  & \textbf{0.9375} (0.9553, 0.0189) 	&	0.7300 (0.8295, 0.0390) 	&	0.8199 (0.8591, 0.0155)  & \textbf{0.9278} (0.9339, 0.0056) 	\\[1em]
4	&	GRU	&	0.9179 (0.9232, 0.0025)	&	\textbf{0.9428} (0.9486, 0.0048)	&	0.7151 (0.7354, 0.0114)	&	0.8132 (0.8272, 0.0068)	&	0.8845 (0.9113, 0.0142)	\\
&	LSTM	&	0.9138 (0.9242, 0.0049)	&	0.9185 (0.9496, 0.0297)	&	0.7208 (0.8015, 0.0252)	&	0.8070 (0.8409, 0.0109)	&	0.9044 (0.9203, 0.0174)	\\
&	LSTM-Attention	&	0.9080 (0.9144, 0.0020)	&	0.8791 (0.9022, 0.0136)	&	0.7332 (0.7665, 0.0162)	&	0.7993 (0.8174, 0.0058)	&	0.9058 (0.9154, 0.0097)	\\
&	LSTM-Transformer	&	0.9134 (0.9181, 0.0044)	&	0.9277 (0.9490, 0.0271)	&	0.7104 (0.7837, 0.0284)	&	0.8038 (0.8172, 0.0122)	&	0.8997 (0.9169, 0.0109)	\\
&	\textbf{GRU-KAN} & \textbf{0.9209} (0.9269, 0.0040) & 0.9351 (0.9440, 0.0071) &	\textbf{0.7346} (0.7592, 0.0171) & \textbf{0.8227} (0.8373, 0.0105) & \textbf{0.9095} (0.9169, 0.0090) 	\\
&	\textbf{LSTM-KAN}	&	0.9182 (0.9278, 0.0045) 	&	0.9263 (0.9476, 0.0246) 	&	0.7320 (0.7744, 0.0235) 	&	0.8172 (0.8396, 0.0110) 	&	0.9087 (0.9184, 0.0213) 	\\[1em]
5	&	GRU	&	0.8979 (0.9112, 0.0350)	&	0.8781 (0.9394, 0.2070)	&	0.6445 (0.6983, 0.1523)	&	0.7432 (0.7972, 0.1752)	&	0.8596 (0.9049, 0.0308)	\\
&	LSTM	&	0.9030 (0.9150, 0.0058)	&	0.8996 (0.9390, 0.0281)	&	0.6702 (0.7302, 0.0194)	&	0.7681 (0.8052, 0.0128)	&	0.8844 (0.8957, 0.0124)	\\
&	LSTM-Attention	&	0.8825 (0.9001, 0.0454)	&	0.7806 (0.8843, 0.2671)	&	0.6259 (0.7188, 0.2150)	&	0.6944 (0.7825, 0.2377)	&	0.8882 (0.9025, 0.0276)	\\
&	LSTM-Transformer	&	0.8851 (0.9116, 0.0468)	&	0.8231 (0.9424, 0.2832)	&	0.5979 (0.7135, 0.2077)	&	0.6915 (0.8007, 0.2374)	&	0.8709 (0.8963, 0.0264)	\\
& \textbf{GRU-KAN} & \textbf{0.9104} (0.9152, 0.0029) & \textbf{0.9111} (0.9392, 0.0227) & \textbf{0.7117} (0.7337, 0.0134) & \textbf{0.7989} (0.8076, 0.0050) & \textbf{0.8949} (0.9079, 0.0121)\\
&	\textbf{LSTM-KAN}	&	0.9074 (0.9125, 0.0037) 	&	0.9110 (0.9350, 0.0325) 	&	0.6994 (0.7387, 0.0219) 	&	0.7906 (0.8016, 0.0067) 	&	0.8797 (0.9064, 0.0338) 	\\[1em]
6	&	GRU	&	0.8792 (0.8909, 0.0305)	&	0.8559 (0.9136, 0.2018)	&	0.5807 (0.6265, 0.1370)	&	0.6918 (0.7417, 0.1630)	&	0.8164 (0.8860, 0.0282)	\\
&	LSTM	&	0.8753 (0.9112, 0.0106)	&	0.8904 (0.9231, 0.0231)	&	0.6090 (0.7617, 0.0442)	&	0.7233 (0.8286, 0.0309)	&	0.8529 (0.9006, 0.0458)	\\
&	LSTM-Attention	&	0.8655 (0.8906, 0.0396)	&	0.8643 (1.0000, 0.0474)	&	0.5619 (0.6770, 0.1928)	&	0.6476 (0.7557, 0.2214)	&	0.8585 (0.8994, 0.0343) 	\\
&	LSTM-Transformer	&	0.8549 (0.8918, 0.0456)	&	0.7679 (0.9119, 0.3312)	&	0.4703 (0.6289, 0.2048)	&	0.5828 (0.7440, 0.2520)	&	0.8439 (0.8862, 0.0366)	\\
&	\textbf{GRU-KAN} & \textbf{0.8928} (0.9030, 0.0044)  &	\textbf{0.9105} (0.9182, 0.0106) 	&	\textbf{0.6336} (0.6722, 0.0158)  &	\textbf{0.7471} (0.7761, 0.0120) & 0.8623 (0.8946, 0.0294)	\\
&	\textbf{LSTM-KAN}	&	0.8877 (0.8928, 0.0036) 	&	0.9072 (0.9170, 0.0206) 	&	0.6140 (0.6280, 0.0099) 	&	0.7322 (0.7455, 0.0072) 	&	\textbf{0.8839} (0.8974, 0.0187)	\\[1em]
7	&	GRU	&	0.8259 (0.8903, 0.0704)	&	0.4955 (0.9068, 0.4599)	&	0.3409 (0.6298, 0.3165)	&	0.4039 (0.7416, 0.3749)	&	0.8118 (0.8527, 0.0334)	\\
&	LSTM	&	0.8846 (0.9008, 0.0046)	&	0.9056 (0.9165, 0.0112)	&	0.6011 (0.6765, 0.0214)	&	0.7223 (0.7733, 0.0142)	&	0.8499 (0.8735, 0.0272)	\\
&	LSTM-Attention	&	0.8672 (0.8920, 0.0397)	&	0.8376 (1.0000, 0.2004)	&	0.5485 (0.7182, 0.1886)	&	0.6465 (0.7688, 0.2167)	&	0.8238 (0.8595, 0.0214)	\\
&	LSTM-Transformer	&	0.8455 (0.8887, 0.0506)	&	0.7144 (0.9281, 0.3674)	&	0.4365 (0.7013, 0.2354)	&	0.5380 (0.7554, 0.2812)	&	0.8422 (0.8754, 0.0283)	\\
&	\textbf{GRU-KAN} & \textbf{0.8947} (0.9062, 0.0074) & 0.8982 (0.9130, 0.0249) & \textbf{0.6533} (0.7010, 0.0264) &	\textbf{0.7560} (0.7889, 0.0187) & \textbf{0.8604} (0.8818, 0.0190) \\
&	\textbf{LSTM-KAN}	&	0.8897 (0.9054, 0.0070) 	&	\textbf{0.9140} (0.9216, 0.0103) 	&	0.6172 (0.7112, 0.0396) 	&	0.7360 (0.7899, 0.0237) 	&	0.8579 (0.8834, 0.0200) 	\\[1em]
8	&	GRU	&	0.7665 (0.8796, 0.0403)	&	0.2284 (1.0000, 0.4070)	&	0.0765 (0.6027, 0.1874)	&	0.0982 (0.7117, 0.2285)	&	0.7744 (0.8513, 0.0302)	\\
&	LSTM	&	0.8769 (0.8989, 0.0065)	&	0.8732 (0.8963, 0.0233)	&	0.5948 (0.6851, 0.0300)	&	0.7069 (0.7720, 0.0193)	&	0.8446 (0.8741, 0.0165)	\\
&	LSTM-Attention	&	0.8516 (0.8793, 0.0387)	&	0.8221 (1.0000, 0.1983)	&	0.4874 (0.6775, 0.1851)	&	0.5939 (0.7346, 0.2144)	&	0.8376 (0.8624, 0.0353)	\\
&	LSTM-Transformer	&	0.8278 (0.8680, 0.0488)	&	0.6641 (0.9396, 0.3943)	&	0.3599 (0.5587, 0.2277)	&	0.4615 (0.6756, 0.2835)	&	0.8220 (0.8619, 0.0313)	\\
& \textbf{GRU-KAN} & \textbf{0.8860} (0.8988, 0.0082) & 0.8707 (0.8973, 0.0320) &	\textbf{0.6405} (0.6927, 0.0332) &	\textbf{0.7372} (0.7736, 0.0211) 	&	\textbf{0.8570} (0.8720, 0.0177) 	\\
&	\textbf{LSTM-KAN}	&	0.8796 (0.8930, 0.0086) 	&	\textbf{0.8752} (0.9034, 0.0423) 	&	0.6072 (0.6494, 0.0312) 	&	0.7158 (0.7521, 0.0197) 	&	0.8548 (0.8737, 0.0211) 	\\
\bottomrule
\end{tabular}
\label{tab:exp2}
\end{sidewaystable}

\begin{sidewaystable}[!ht]
\tiny
\centering
\caption{Results of model performance based on 20 trials using different sample sizes}
\begin{tabular}{llccccc}
\toprule
Sample size	& Model & Accuracy	(Best, Std) & Precision (Best, Std)	& Recall (Best, Std) & F1	(Best, Std) & AUC (Best, Std) \\ \midrule
500000	&	GRU	&	0.8769 (0.9825, 0.1102)	&	0.5911 (1.0000, 0.4951)	&	0.5157 (0.9484, 0.4485)	&	0.5453 (0.9644, 0.4643)	&	0.9056 (0.9775, 0.0766)	\\
	&	LSTM	&	0.9577 (0.9603, 0.0024) 	&	0.9594 (0.9796, 0.0184) 	&	0.8536 (0.8765, 0.0148) 	&	0.9034 (0.9169, 0.0081) 	&	0.9637 (0.9721, 0.0066) 	\\
	&	LSTM-Attention	&	0.8019 (0.9833, 0.0912)	&	0.4742 (1.0000, 0.4877)	&	0.2121 (0.9541, 0.3721)	&	0.2931 (0.9662, 0.3878)	&	0.8516 (0.9773, 0.1015)	\\
	&	LSTM-Transformer	&	0.9001 (0.9544, 0.0811)	&	0.9858 (1.0000, 0.0105)	&	0.6119 (0.8378, 0.3314)	&	0.7551 (0.9018, 0.3551)	&	0.9001 (0.9142, 0.0047)	\\
	&	\textbf{GRU-KAN}	&	\textbf{0.9707} (0.9851, 0.0137)	&	\textbf{0.9874} (1.0000, 0.0059)	&	\textbf{0.8946} (0.9554, 0.0581)	&	\textbf{0.9376} (0.9698, 0.0326)	&	0.9681 (0.9881, 0.0225)	\\
	&	\textbf{LSTM-KAN}	&	0.9638 (0.9851, 0.0264)	&	0.9831 (0.9939, 0.0071)	&	0.8705 (0.9604, 0.1098)	&	0.9190 (0.9700, 0.0738)	&	\textbf{0.9744} (0.9900, 0.0175)	\\[1em]
1000000	&	GRU	&	0.8885 (0.9827, 0.1160)	&	0.5857 (0.9776, 0.4907)	&	0.5679 (0.9536, 0.4758)	&	0.5767 (0.9651, 0.4831)	&	0.9094 (0.9783, 0.0882)	\\
	&	LSTM	&	0.9671 (0.9825, 0.0126)	&	0.9612 (0.9820, 0.0114)	&	0.8954 (0.9570, 0.0528)	&	0.9271 (0.9568, 0.0286)	&	0.9697 (0.9758, 0.0046)	\\
	&	LSTM-Attention	&	0.8465 (0.9714, 0.1025)	&	0.4694 (0.9769, 0.4818)	&	0.4144 (0.9458, 0.4417)	&	0.4343 (0.9427, 0.4524)	&	0.9046 (0.9860, 0.0622)	\\
	&	LSTM-Transformer	&	0.9078 (0.9788, 0.0843)	&	0.8876 (1.0000, 0.3036)	&	0.6416 (0.9331, 0.3427)	&	0.7448 (0.9565, 0.3658)	&	0.9171 (0.9839, 0.0414)	\\
	&	\textbf{GRU-KAN}	&	0.9767 (0.9854, 0.0097)	&	\textbf{0.9803} (0.9912, 0.0116)	&	0.9257 (0.9541, 0.0358)	&	0.9519 (0.9703, 0.0210)	&	0.9757 (0.9874, 0.0173)	\\
	&	\textbf{LSTM-KAN}	&	\textbf{0.9811} (0.9854, 0.0053)	&	0.9773 (0.9870, 0.0069)	&	\textbf{0.9462} (0.9582, 0.0199)	&	\textbf{0.9614} (0.9704, 0.0114)	&	\textbf{0.9845} (0.9898, 0.0075)	\\[1em]
1500000	&	GRU	&	0.9456 (0.9625, 0.0149)	&	0.8783 (0.9805, 0.3004)	&	0.8329 (0.9475, 0.2853)	&	0.8550 (0.9602, 0.2925)	&	0.9376 (0.9697, 0.0308)	\\
	&	LSTM	&	0.9721 (0.9840, 0.0014)	&	0.9622 (0.9706, 0.0053)	&	0.9349 (0.9382, 0.0027)	&	0.9484 (0.9531, 0.0089)	&	0.9767 (0.9788, 0.0014)	\\
	&	LSTM-Attention	&	0.9143 (0.9618, 0.0571)	&	0.9340 (1.0000, 0.2203)	&	0.6706 (0.8706, 0.2365)	&	0.7807 (0.9193, 0.2326)	&	0.9153 (0.9284, 0.0154)	\\
	&	LSTM-Transformer	&	0.9531 (0.9804, 0.0696)	&	0.9510 (0.9782, 0.0174)	&	0.8324 (0.8959, 0.1170)	&	0.8878 (0.9217, 0.0810)	&	0.9493 (0.9746, 0.0557)	\\
	&	\textbf{GRU-KAN}	&	0.9798 (0.9841, 0.0124)	&	0.9727 (0.9873, 0.0425)	&	\textbf{0.9482} (0.9616, 0.0091)	&	0.9597 (0.9678, 0.0216)	&	0.9835 (0.9868, 0.0033)	\\
	&	\textbf{LSTM-KAN}	&	\textbf{0.9823} (0.9835, 0.0010)	&	\textbf{0.9817} (0.9848, 0.0035)	&	0.9468 (0.9577, 0.0062)	&	\textbf{0.9639} (0.9665, 0.0021)	&	\textbf{0.9864} (0.9882, 0.0011)	\\[1em]
2000000	&	GRU	&	0.9406 (0.9659, 0.0245)	&	0.8948 (0.8972, 0.0135)	&	0.8568 (0.8844, 0.0377)	&	0.8754 (0.8897, 0.0078)	&	0.9461 (0.9708, 0.0326)	\\
	&	LSTM	&	0.9718 (0.9834, 0.0110)	&	0.9613 (0.9790, 0.0152)	&	0.9456 (0.9511, 0.0087)	&	0.9534 (0.9563, 0.0019)	&	0.9858 (0.9885, 0.0019)	\\
	&	LSTM-Attention	&	0.9437 (0.9659, 0.0424)	&	0.9702 (0.9890, 0.0083)	&	0.7999 (0.8883, 0.1748)	&	0.8769 (0.9287, 0.1595)	&	0.9487 (0.9747, 0.0324)	\\
	&	LSTM-Transformer	&	0.9449 (0.9666, 0.0181)	&	0.9675 (0.9905, 0.0255)	&	0.8075 (0.8935, 0.0741)	&	0.8782 (0.9300, 0.0444)	&	0.9320 (0.9706, 0.0345)	\\
	&	\textbf{GRU-KAN}	&	\textbf{0.9823} (0.9830, 0.0007)	&	\textbf{0.9747} (0.9787, 0.0044)	&	0.9539 (0.9596, 0.0024)	&	\textbf{0.9642} (0.9656, 0.0013)	&	0.9853 (0.9881, 0.0023)	\\
	&	\textbf{LSTM-KAN}	&	0.9821 (0.9837, 0.0010)	&	0.9732 (0.9807, 0.0058)	&	\textbf{0.9547} (0.9596, 0.0032)	&	0.9638 (0.9670, 0.0018)	&	\textbf{0.9865} (0.9894, 0.0013)	\\[1em]
3000000	&	GRU	&	0.9565 (0.9669, 0.0505)	&	0.9246 (0.9612, 0.0180)	&	0.8797 (0.8908, 0.0083)	&	0.9016 (0.9207, 0.0643)	&	0.9334 (0.9663, 0.0459)	\\
	&	LSTM	&	0.9764 (0.9828, 0.0012)	&	0.9684 (0.9785, 0.0058)	&	0.9504 (0.9569, 0.0018)	&	0.9593 (0.9651, 0.0023)	&	0.9847 (0.9865, 0.0020)	\\
	&	LSTM-Attention	&	0.9616 (0.9791, 0.0499)	&	0.9182 (0.9706, 0.1619)	&	0.8986 (0.9594, 0.2117)	&	0.9083 (0.9575, 0.1915)	&	0.9773 (0.9856, 0.0266)	\\
	&	LSTM-Transformer	&	0.9396 (0.9654, 0.0496)	&	0.9126 (0.9868, 0.2215)	&	0.7893 (0.9014, 0.2077)	&	0.8440 (0.9277, 0.2101)	&	0.9336 (0.9647, 0.0402)	\\
	&	\textbf{GRU-KAN}	&	\textbf{0.9838} (0.9847, 0.0017)	&	\textbf{0.9765} (0.9814, 0.0100)	&	\textbf{0.9524} (0.9581, 0.0041)	&	\textbf{0.9643} (0.9651, 0.0032)	&	\textbf{0.9859} (0.9872, 0.0022)	\\
	&	\textbf{LSTM-KAN}	&	0.9807 (0.9829, 0.0026)	&	0.9714 (0.9814, 0.0138)	&	0.9511 (0.9581, 0.0045)	&	0.9611 (0.9652, 0.0049)	&	0.9839 (0.9861, 0.0020)	\\[1em]
5000000	&	GRU	&	0.9793 (0.9817, 0.0023)	&	0.9585 (0.9841, 0.0095)	&	0.9211 (0.9517, 0.0109)	&	0.9394 (0.9656, 0.0048)	&	0.9736 (0.9785, 0.0026)	\\
	&	LSTM	&	0.9814 (0.9828, 0.0012)	&	0.9716 (0.9785, 0.0058)	&	0.9534 (0.9569, 0.0018)	&	0.9624 (0.9664, 0.0025)	&	0.9847 (0.9865, 0.0020)	\\
	&	LSTM-Attention	&	0.9776 (0.9812, 0.0027)	&	0.9548 (0.9697, 0.0116)	&	0.9157 (0.9619, 0.2040)	&	0.9348 (0.9558, 0.0059)	&	0.9754 (0.9773, 0.0018) 	\\
	&	LSTM-Transformer	&	0.9808 (0.9836, 0.0043)	&	0.9753 (0.9799, 0.0031)	&	0.9368 (0.9497, 0.0425)	&	0.9557 (0.9594, 0.0155)	&	0.9829 (0.9846, 0.0012)	\\
	&	\textbf{GRU-KAN}	&	\textbf{0.9854} (0.9869, 0.0015)	&	\textbf{0.9790} (0.9808, 0.0009)	&	\textbf{0.9588} (0.9658, 0.0052)	&	\textbf{0.9688} (0.9705, 0.0031)	&	\textbf{0.9858} (0.9885, 0.0019)	\\
	&	\textbf{LSTM-KAN}	&	0.9822 (0.9825, 0.0005)	&	0.9762 (0.9833, 0.0062)	&	0.9491 (0.9525, 0.0027)	&	0.9625 (0.9645, 0.0012)	&	0.9837 (0.9856, 0.0016)	\\
\bottomrule
\end{tabular}
\label{tab:exp1}
\end{sidewaystable}

\begin{sidewaystable}[!ht]
\tiny
\centering
\caption{Results of model performance based on 20 trials using different cohorts of training and test sets}
\begin{tabular}{llccccc}
\toprule
Training$\to$Test	&	Model	&	Accuracy (Best, Std)	&	Precision (Best, Std)	&	Recall (Best, Std)	&	F1 (Best, Std)	&	AUC (Best, Std)	\\ \midrule
2018$\to$2019	&	GRU	&	0.8217 (0.9147, 0.0649)	&	0.7656 (0.9103, 0.1211)	&	0.3491 (0.7707, 0.2881)	&	0.4275 (0.8188, 0.3019)	&	0.8231 (0.8995, 0.0676)	\\
	&	LSTM	&	0.8845 (0.9075, 0.0249)	&	0.8155 (0.9038, 0.0961)	&	0.7222 (0.8082, 0.0540)	&	0.7594 (0.8027, 0.0330)	&	0.8624 (0.8933, 0.0142)	\\
	&	LSTM-Attention	&	0.7505 (0.7532, 0.0010)	&	0.4549 (1.0000, 0.2703)	&	0.0211 (0.0332, 0.0119)	&	0.0401 (0.0626, 0.0224)	&	0.6978 (0.8133, 0.1223)	\\
	&	LSTM-Transformer	&	0.7795 (0.8994, 0.0586)	&	0.2295 (0.8848, 0.3610)	&	0.1405 (0.7050, 0.2718)	&	0.1607 (0.7780, 0.3042)	&	0.6669 (0.8567, 0.1357)	\\
	&	\textbf{GRU-KAN}	&	\textbf{0.9064} (0.9113, 0.0030)	&	0.8615 (0.8775, 0.0136)	&	\textbf{0.7460} (0.7801, 0.0182)	&	\textbf{0.7993} (0.8123, 0.0077)	&	\textbf{0.8768} (0.9020, 0.0198)	\\
	&	\textbf{LSTM-KAN}	&	0.9007 (0.9090, 0.0064)	&	\textbf{0.8835} (0.8960, 0.0117)	&	0.6949 (0.7298, 0.0294)	&	0.7775 (0.8004, 0.0183)	&	0.8650 (0.8909, 0.0115)	\\[1em]
2018$\to$2020	&	GRU	&	0.8349 (0.9250, 0.0493)	&	\textbf{0.9717} (0.9890, 0.0244)	&	0.3550 (0.7705, 0.2151)	&	0.4837 (0.8370, 0.2038)	&	0.8922 (0.9116, 0.0110)	\\
	&	LSTM	&	0.9135 (0.9313, 0.0086)	&	0.9108 (0.9597, 0.0468)	&	0.7286 (0.7773, 0.0253)	&	0.8084 (0.8463, 0.0154)	&	0.8941 (0.9345, 0.0255)	\\
	&	LSTM-Attention	&	0.7856 (0.9006, 0.0439)	&	0.6777 (1.0000, 0.4443)	&	0.1516 (0.6727, 0.1947)	&	0.2190 (0.7718, 0.2346)	&	0.8562 (0.9014, 0.0627)	\\
	&	LSTM-Transformer	&	0.7597 (0.7892, 0.0148)	&	0.3947 (1.0000, 0.4834)	&	0.0393 (0.1591, 0.0602)	&	0.0694 (0.2740, 0.1056)	&	0.7680 (0.9128, 0.1532)	\\
	&	\textbf{GRU-KAN}	&	\textbf{0.9340} (0.9489, 0.0080)	&	0.9620 (0.9783, 0.0189)	&	\textbf{0.7670} (0.8227, 0.0363)	&	\textbf{0.8528} (0.8894, 0.0208)	&	\textbf{0.9320} (0.9558, 0.0192)	\\
	&	\textbf{LSTM-KAN}	&	0.9205 (0.9369, 0.0090)	&	0.9257 (0.9637, 0.0303)	&	0.7443 (0.8682, 0.0610)	&	0.8226 (0.8731, 0.0274)	&	0.9072 (0.9639, 0.0386)	\\[1em]
2018$\to$2021	&	GRU	&	0.8209 (0.8705, 0.0273)	&	\textbf{1.0000} (1.0000, 0.0000)	&	0.2836 (0.4818, 0.1092)	&	0.4310 (0.6503, 0.1284)	&	0.8321 (0.8669, 0.0167)	\\
	&	LSTM	&	0.8636 (0.8682, 0.0047)	&	0.9892 (1.0000, 0.0267)	&	0.4600 (0.4818, 0.0153)	&	0.6277 (0.6463, 0.0145)	&	0.8147 (0.8536, 0.0244)	\\
	&	LSTM-Attention	&	0.7802 (0.7977, 0.0080)	&	\textbf{1.0000} (1.0000, 0.0000)	&	0.1209 (0.1909, 0.0320)	&	0.2143 (0.3206, 0.0502)	&	0.8102 (0.8263, 0.0095)	\\
	&	LSTM-Transformer	&	0.7511 (0.7591, 0.0027)	&	0.2000 (1.0000, 0.4000)	&	0.0045 (0.0364, 0.0109)	&	0.0088 (0.0702, 0.0211)	&	0.7669 (0.8388, 0.0595)	\\
	&	\textbf{GRU-KAN}	&	\textbf{0.8793} (0.8864, 0.0050)	&	0.9835 (1.0000, 0.0348)	&	\textbf{0.5273} (0.5545, 0.0195)	&	\textbf{0.6858} (0.7093, 0.0142)	&	\textbf{0.8408} (0.8551, 0.0065)	\\
	&	\textbf{LSTM-KAN}	&	0.8655 (0.8750, 0.0078)	&	0.9719 (1.0000, 0.0399)	&	0.4764 (0.5000, 0.0200)	&	0.6389 (0.6667, 0.0214)	&	0.8304 (0.8728, 0.0269)	\\ [1em]
2018$\to$2022	&	GRU	&	0.7561 (0.7579, 0.0014)	&	\textbf{1.0000} (1.0000, 0.0000)	&	0.0246 (0.0316, 0.0057)	&	0.0479 (0.0612, 0.0108)	&	0.6429 (0.7341, 0.0515)	\\
	&	LSTM	&	0.7504 (0.7535, 0.0011)	&	0.1000 (1.0000, 0.3000)	&	0.0014 (0.0140, 0.0042)	&	0.0028 (0.0277, 0.0083)	&	0.6061 (0.7748, 0.1248)	\\
	&	LSTM-Attention	&	0.7558 (0.7605, 0.0021)	&	\textbf{1.0000} (1.0000, 0.0000)	&	0.0232 (0.0421, 0.0085)	&	0.0451 (0.0808, 0.0161)	&	0.6655 (0.7386, 0.0308)	\\
	&	LSTM-Transformer	&	0.7555 (0.7816, 0.0088)	&	\textbf{1.0000} (1.0000, 0.0000)	&	0.0221 (0.1263, 0.0351)	&	0.0411 (0.2243, 0.0619)	&	0.6799 (0.7025, 0.0151)	\\
	&	\textbf{GRU-KAN}	&	\textbf{0.8132} (0.8412, 0.0202)	&	\textbf{1.0000} (1.0000, 0.0000)	&	0.2526 (0.3649, 0.0809)	&	\textbf{0.3965} (0.5347, 0.1067)	&	0.7512 (0.7786, 0.0369)	\\
	&	\textbf{LSTM-KAN}	&	0.8118 (0.8412, 0.0250)	&	0.9775 (1.0000, 0.0432)	&	\textbf{0.2568} (0.3965, 0.1108)	&	0.3920 (0.5531, 0.1422)	&	\textbf{0.7774} (0.7991, 0.0170)	\\[1em]
2019$\to$2021	&	GRU	&	0.8776 (0.8880, 0.0062)	&	\textbf{1.0000} (1.0000, 0.0000)	&	0.5104 (0.5521, 0.0247)	&	0.6755 (0.7114, 0.0214)	&	0.8625 (0.8800, 0.0084)	\\
	&	LSTM	&	0.8799 (0.8854, 0.0036)	&	\textbf{1.0000} (1.0000, 0.0000)	&	0.5198 (0.5417, 0.0143)	&	0.6839 (0.7027, 0.0124)	&	0.8677 (0.8830, 0.0057)	\\
	&	LSTM-Attention	&	0.8734 (0.8854, 0.0090)	&	\textbf{1.0000} (1.0000, 0.0000)	&	0.4938 (0.5417, 0.0361)	&	0.6603 (0.7027, 0.0327)	&	0.8692 (0.8807, 0.0074)	\\
	&	LSTM-Transformer	&	0.8797 (0.8880, 0.0054)	&	\textbf{1.0000} (1.0000, 0.0000)	&	0.5188 (0.5521, 0.0218)	&	0.6829 (0.7114, 0.0188)	&	0.8609 (0.8877, 0.0237)	\\
	&	\textbf{GRU-KAN}	&	\textbf{0.8846} (0.8984, 0.0053)	&	\textbf{1.0000} (1.0000, 0.0000)	&	\textbf{0.5385} (0.5938, 0.0214)	&	\textbf{0.6998} (0.7451, 0.0176)	&	0.8667 (0.8907, 0.0120)	\\
	&	\textbf{LSTM-KAN}	&	0.8844 (0.8932, 0.0045)	&	\textbf{1.0000} (1.0000, 0.0000)	&	0.5375 (0.5729, 0.0182)	&	0.6990 (0.7285, 0.0151)	&	\textbf{0.8730} (0.8815, 0.0082)	\\[1em]
2019$\to$2022	&	GRU	&	0.8276 (0.8298, 0.0011)	&	\textbf{1.0000} (1.0000, 0.0000)	&	0.3105 (0.3193, 0.0042)	&	0.4739 (0.4840, 0.0049)	&	0.7566 (0.7775, 0.0188)	\\
	&	LSTM	&	0.8332 (0.8404, 0.0037)	&	\textbf{1.0000} (1.0000, 0.0000)	&	0.3330 (0.3614, 0.0148)	&	0.4994 (0.5309, 0.0166)	&	0.7941 (0.8011, 0.0044)	\\
	&	LSTM-Attention	&	0.8312 (0.8368, 0.0034)	&	\textbf{1.0000} (1.0000, 0.0000)	&	0.3249 (0.3474, 0.0135)	&	0.4903 (0.5156, 0.0153)	&	0.7869 (0.7965, 0.0084)	\\
	&	LSTM-Transformer	&	0.8334 (0.8386, 0.0029)	&	\textbf{1.0000} (1.0000, 0.0000)	&	0.3337 (0.3544, 0.0117)	&	0.5003 (0.5233, 0.0131)	&	0.7753 (0.7806, 0.0040)	\\
	&	\textbf{GRU-KAN}	&	\textbf{0.8429} (0.8491, 0.0045)	&	\textbf{1.0000} (1.0000, 0.0000)	&	\textbf{0.3716} (0.3965, 0.0179)	&	\textbf{0.5416} (0.5678, 0.0190)	&	\textbf{0.8020} (0.8111, 0.0084)	\\
	&	\textbf{LSTM-KAN}	&	0.8360 (0.8368, 0.0007)	&	\textbf{1.0000} (1.0000, 0.0000)	&	0.3439 (0.3474, 0.0027)	&	0.5117 (0.5156, 0.0030)	&	0.7942 (0.8001, 0.0047)	\\
\bottomrule
\end{tabular}
\label{tab:exp4}
\end{sidewaystable}

\end{appendices}



\end{document}